\documentclass[sigconf]{acmart}

\AtBeginDocument{%
  }

\setcopyright{acmlicensed}
\copyrightyear{2026}
\acmYear{2026}
\acmDOI{XXXXXXX.XXXXXXX}

\acmISBN{978-1-4503-XXXX-X/2018/06}

\usepackage{algorithm}
\usepackage{algpseudocode}
\usepackage{amsmath} 

\algrenewcommand{\algorithmiccomment}[1]{\hfill $\triangleright$ #1}

\usepackage{amsmath}
\usepackage{ulem}
\usepackage{bm}
\usepackage{enumitem}
\usepackage{multirow}
\usepackage{graphicx}
\usepackage{subcaption}
\usepackage[export]{adjustbox}

\usepackage{pifont} 
\usepackage{geometry}
\usepackage{diagbox}
\definecolor{myBlue}{HTML}{3366CC}
\definecolor{myGreen}{HTML}{669933}

\definecolor{lightgreen}{RGB}{230, 255, 230}
\definecolor{darkgreen}{RGB}{0, 100, 0}
\newcommand{\cmark}{\textcolor{darkgreen}{\ding{51}}} 

\newcommand{\specialcell}[2][c]{%
	\begin{tabular}[#1]{@{}c@{}}#2\end{tabular}}

\renewcommand\footnotetextcopyrightpermission[1]{} 
\begin{document}

\title{\textsc{DDIAgents}: Mechanism-Conditioned Context Flow for Drug-Drug Interaction Prediction}

\author{Zhenqian Shen}
\email{szg22@mails.tsinghua.edu.cn}
\affiliation{%
  \department{Department of Electronic Engineering, }
  \institution{Tsinghua University}
  \city{Beijing}
  \country{China}
}

\author{Yu Liu}
\email{yu.liu@eng.ox.ac.uk}
\affiliation{%
  \department{Institute of Biomedical Engineering, }
  \institution{University of Oxford}
  \city{Oxford}
  \country{United Kingdom}
}

\author{Xiaoyi Fu}
\email{xiaoyif@alumni.cmu.edu}
\affiliation{%
  \department{Division of Emerging Interdisciplinary Areas, }
  \institution{Hong Kong University of Science and Technology}
  \city{Hong Kong}
  \country{China}
}

\author{Quanming Yao}
\email{qyaoaa@tsinghua.edu.cn}
\affiliation{%
  \department{Department of Electronic Engineering}
  \institution{Tsinghua University}
  \city{Beijing}
  \country{China}
}
\authornote{Corresponding author}






\begin{abstract}
Drug-drug interaction~(DDI) prediction is essential for medication safety, yet it requires reasoning over heterogeneous biomedical evidence whose relevance changes across interaction mechanisms.
We propose DDIAgents, a mechanism-conditioned multi-agent framework that performs DDI prediction through dynamic knowledge orchestration. 
Given a drug pair, a planner agent instantiates specialized expert agents, routes mechanism-relevant knowledge sources to each agent, and aggregates their analyses through a conclusion agent. 
By adapting context flow to the inferred interaction mechanism, DDIAgents reduces irrelevant information, supports complementary expert reasoning, and produces interpretable agent-level rationales. 
Extensive experiments on realistic DDI prediction benchmarks show that DDIAgents consistently outperforms existing feature-based, graph-based, LLM-based, and agent-based baselines. 
Beyond prediction performance, DDIAgents demonstrates how multi-agent systems can organize heterogeneous scientific knowledge for adaptive and interpretable AI4Science reasoning.
\end{abstract}





\maketitle

\section{Introduction}

Modern drug discovery is defined by prohibitive costs and protracted timelines, frequently taking 10--15 years and over 2 billion dollars per approved drug~\citep{dimasi2016innovation, prasad2017high}. 
It is therefore crucial to identify drug-drug interactions (DDIs) that could lead to adverse drug reactions and serious health risks for patients. 
However, DDI prediction remains a challenging scientific problem due to intricate biomedical mechanisms, including altered pharmacokinetics and pharmacodynamics~\citep{palleria2013pharmacokinetic}, alongside clinical treatment factors~\citep{roberts2018mechanisms}.  
Moreover, candidate DDI verification is expensive and time-consuming, requiring extensive experiments from preclinical studies to large-scale clinical trials.
This setting reflects a broader data-efficient AI4Science challenge: direct labels and experimental feedback are costly, so models must make better use of prior knowledge, task structure, and limited feedback rather than relying only on scaling supervision~\citep{wang2020generalizing,wang2026beyond}. 
In this work, we focus on a complementary agentic question: how should a reasoning system decide which biomedical evidence each expert should use when the underlying interaction mechanism changes across drug pairs?

To accelerate DDI discovery, a variety of computational methods have been developed for prediction~\citep{qiu2021comprehensive, lin2023comprehensive}. 
These methods broadly fall into three categories: feature-based, graph-based, and large language model (LLM)-based approaches. 
Feature-based methods~\citep{ryu2018deep, liu2022predict} typically encode drugs with molecular features and train neural networks for interaction prediction. 
Graph-based methods broadly utilize graph embedding methods, graph neural networks (GNNs) or graph transformers to extract predictive features from diverse data sources, ranging from individual drug molecular structures~\citep{nyamabo2021ssi, yang2022learning} to complex biomedical networks~\citep{zitnik2018modeling, zhang2023emerging}.
LLM-based methods primarily utilize the powerful reasoning capabilities based on textual descriptions of drugs and interactions~\citep{zhu2023learning, xu2024ddi}.

Despite these advances, DDI prediction is not merely a single-pattern learning problem; it is a mechanism-conditioned knowledge orchestration problem. 
First, different drug pairs can be governed by \textbf{diverse mechanisms}, from physiological modulation and drug--target binding events to chemically mediated reactions. 
Second, the decisive evidence exhibits \textbf{knowledge heterogeneity}, spanning natural-language descriptions, structured biomedical relations, and molecular substructures. 
A metabolism-driven interaction may require enzyme or transporter evidence, whereas a target-overlap interaction may depend more on pathway or adverse-effect evidence. 
However, most existing approaches remain inflexible: they follow fixed reasoning patterns and rely on static evidence integration, which can introduce irrelevant evidence or miss the modality that is decisive for a specific mechanism.

Meanwhile, the recent emergence of LLM-based multi-agent systems in other scientific domains has become a promising solution for providing adaptable mechanism analysis of DDIs. 
Analogous to a multidisciplinary research team, multi-agent frameworks coordinate multiple role-specialized ``expert'' agents to solve complex problems~\citep{boiko2023autonomous, swanson2025virtual}. 
This perspective also aligns with data-efficient agentic learning, where performance is improved by organizing roles, feedback, and interactions rather than simply increasing labeled data or model scale~\citep{wang2026beyond}. 
In such systems, the reasoning performance critically depends on the \textbf{context flow}, namely how knowledge sources are routed to the agents that need it. 
However, existing multi-agent systems often rely on a fixed context flow, where agents are restricted to a predefined knowledge scope.
This rigidity forces expert agents to handle a fixed scope of knowledge regardless of the specific task, which can introduce irrelevant information into the reasoning process and obscure the critical insights.

To address the challenges above, we introduce DDIAgents, a mechanism-conditioned multi-agent framework for DDI prediction that leverages dynamic context flow for specialized DDI analysis. 
Specifically, DDIAgents conducts the following three stages iteratively. 
Firstly, expert agent instantiation gathers a team of specialized expert agents. 
Secondly, dynamic context flow routes suitable knowledge sources to each expert agent according to the mechanism-specific evidence needs of the current drug pair. 
Finally, expert agents generate domain-specific analyses that are synthesized by a conclusion agent to either yield a final prediction or provide strategic guidance for iterative refinement.
Experiments on two benchmark datasets demonstrate the superior performance of DDIAgents compared to existing methods. 
Furthermore, the agent-based analysis demonstrates that DDIAgents yields interpretable mechanistic explanations of predicted DDIs.
The main contributions are summarized as follows:
\begin{itemize}[leftmargin=*]
\item We formulate DDI prediction as a mechanism-conditioned knowledge orchestration problem and introduce DDIAgents, a multi-agent framework that explicitly accounts for diverse mechanisms and heterogeneous biomedical evidence.
\item We design a dynamic context flow mechanism that strategically routes heterogeneous knowledge sources to specialized agents, thereby enhancing knowledge utility and reducing noisy information in the DDI reasoning process.
\item We perform extensive experiments on two benchmark datasets, demonstrating that DDIAgents consistently outperforms existing feature-based, graph-based, LLM-based, and agent-based approaches. Comprehensive case studies further show that DDIAgents provides interpretable agent-level insights into the underlying mechanisms of DDIs. 
\end{itemize}

\section{Related Works}

\subsection{Drug-Drug Interaction Prediction}

DDI prediction has recently been a critical topic in computational pharmacology, aiming to identify potential drug interactions that may lead to adverse effects or altered therapeutic efficacy~\citep{han2022review, lin2023comprehensive}. 
Existing approaches are commonly grouped into feature-based, graph-based, and LLM-based methods.

Feature-based methods encode drugs with molecular descriptors and train supervised predictors~\citep{rogers2010extended, ryu2018deep, liu2022predict, xie2025predicting}. 
They work well when interactions are driven by chemical properties, but often struggle to incorporate broader biological mechanisms.
Graph-based methods exploit structured relations from molecular graphs and biomedical networks through knowledge graph embedding methods~\cite{yao2022effective}, GNNs~\citep{zitnik2018modeling, zhang2023emerging} or heterogeneous graph transformers~\citep{su2024dual}. 
While they better capture relational evidence, their pipelines and knowledge fusion are typically fixed, limiting adaptability when the decisive evidence varies across drug pairs.
Recently, LLM-based methods prompt models with drug or interaction descriptions~\citep{zhu2023learning,xu2024ddi, raveendran2025pharmamap}, retrieve graph-path evidence~\citep{abdullahi2025k}, or apply case-based reasoning to reuse historical DDI cases~\citep{liu2025case}. 
In particular, CBR-DDI shows that historical cases can provide useful reasoning examples for LLM-based DDI prediction. 
This case-reuse direction is complementary to DDIAgents: our focus is not to build a historical case repository, but to decide, for each drug pair and mechanism hypothesis, how heterogeneous biomedical evidence should be routed to specialized agents. 
Thus, DDIAgents targets mechanism-conditioned knowledge orchestration rather than case-based rationale transfer.

Overall, existing DDI prediction methods provide complementary strengths, but most still follow uniform reasoning patterns and static knowledge utilization. 
This rigidity is a key limitation for DDI prediction, where different drug pairs require different explanatory routes and different forms of supporting evidence.

\subsection{Multi-Agent Reasoning for Scientific Tasks}

Powered by strong natural language understanding and generation, LLMs have recently enabled a new paradigm for tackling complex scientific tasks via multi-agent collaboration~\citep{boiko2023autonomous, kim2024mdagents, swanson2025virtual}. 
Compared with conventional machine learning pipelines, multi-agent systems decompose a task into role-specialized agents, each equipped with distinct domain expertise and responsibilities.
This design principle is closely related to data-efficient agentic learning, which studies how agent systems can adapt under limited supervision, limited feedback, and costly interactions~\citep{wang2026beyond}.

Coscientist~\citep{boiko2023autonomous} is an early framework that formulates an end-to-end workflow for autonomous chemical research, coordinating agents such as planner agent, document searchers and experiment executors. 
SciAgents~\citep{ghafarollahi2025sciagents} builds a materials-science knowledge graph and organizes a team to generate hypotheses by following evidence paths extracted from the knowledge graph. 
Virtual Lab~\citep{swanson2025virtual} instantiates a group of LLM scientist agents for interdisciplinary research and has been applied to novel nanobody design. 
Related directions also explore multi-agent systems for clinical decision-making, assembling expert agents with distinct specialties~\citep{tang2024medagents, kim2024mdagents}.

Despite these advances, many multi-agent systems rely on fixed agent configurations and static information-access patterns. 
In scientific problems where different sub-problems require different knowledge sources, the way information is allocated across agents can substantially influence reasoning quality and interpretability.

\begin{figure*}[t]
	\centering
	\includegraphics[width=0.97\textwidth]{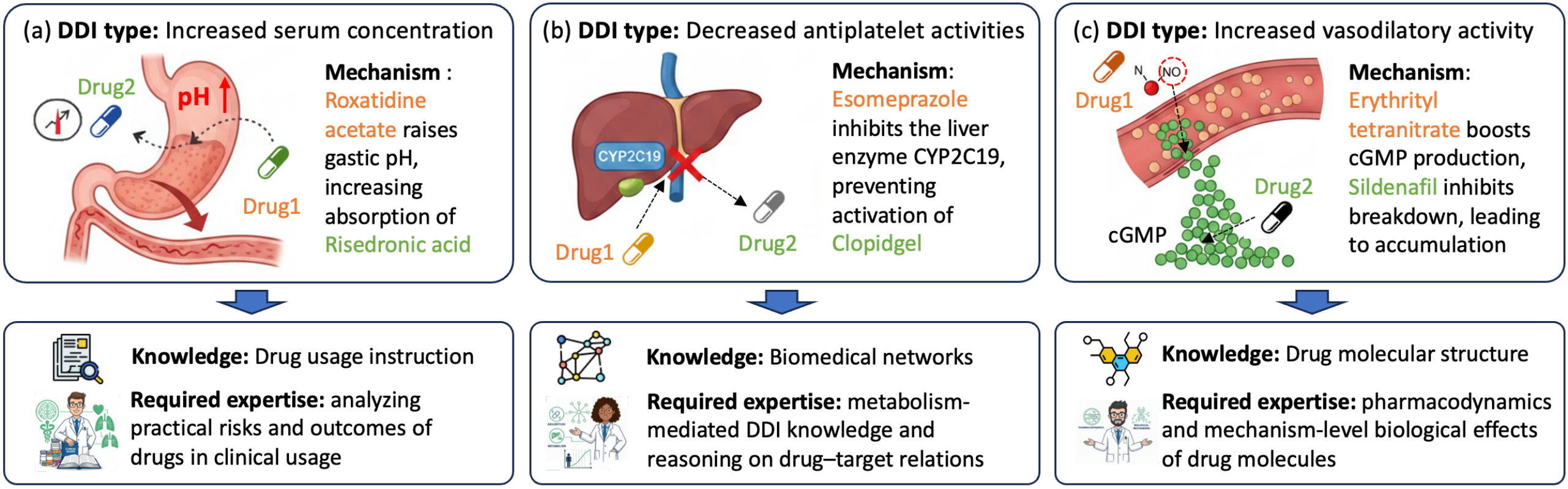}
  \vspace{-5px}
	\caption{Three DDI examples with different mechanisms: (a) pharmacokinetic change, (b) pharmacodynamic interference, (c) chemically mediated synergy. The prediction of these DDIs requires different knowledge and expertise. }
  \vspace{-5px}
	\label{fig:limit}
\end{figure*}

\section{Motivation}

\subsection{DDI Prediction as a Mechanism-conditioned Reasoning Problem}

Drug--drug interaction (DDI) prediction is fundamental to medication safety, yet it remains challenging in realistic settings because 
\textbf{it requires mechanism-conditioned reasoning rather than single-pattern learning}.
Clinically meaningful DDIs arise from different mechanisms, and the correct interaction type for each drug pair depends on identifying the governing mechanism, rather than applying a universal decision rule.

Figure~\ref{fig:limit} highlights that clinically meaningful DDIs can be driven by diverse mechanisms, including pharmacokinetic change, pharmacodynamic interference and chemically mediated synergy. 
These mechanisms correspond to qualitatively different causal explanations and therefore require different reasoning paths. \textbf{Effective DDI prediction thus requires selecting an appropriate explanatory route before determining the interaction type.}

A closely related challenge is that \textbf{the decisive evidence is heterogeneous, mechanism-conditioned, and demands complementary expert perspectives}. 
No single modality or fixed expert viewpoint is uniformly informative across all DDIs. 
When knowledge is aggregated without expert-guided selection, critical evidence can be diluted by irrelevant context and cross-perspective distraction, ultimately undermining reliable prediction. 

\subsection{Limitations of Static Reasoning and the Need for Dynamic Context Flow}

Despite substantial progress, most existing DDI prediction methods implicitly assume static reasoning. 
Feature-based and graph-based models rely on fixed pipelines and static feature fusion, while recent LLM-based methods typically employ fixed prompts and retrieval scopes. As a result, \textbf{they reuse the same reasoning pattern and evidence across drug pairs}, which can either omit the modality most relevant to the governing mechanism or overwhelm the model with irrelevant context. 

Multi-agent reasoning provides a natural abstraction for incorporating multiple expert perspectives, but \textbf{naive multi-agent designs remain insufficient}. 
Most existing multi-agent systems for scientific tasks adopt a fixed context flow, i.e., heterogeneous knowledge is routed to expert agents according to a predetermined scheme.
This rigidity constrains experts to static information scopes and prevents adaptation when mechanisms vary across instances.
\textbf{Therefore, the primary bottleneck is not agent collaboration itself, but how knowledge is allocated to agents.}

These observations lead to a key insight: \textbf{mechanism-selective biomedical reasoning requires instance-adaptive context flow}. 
For each drug pair, the system should dynamically determine which experts to instantiate, which types of knowledge each expert should access, and whether these choices need to be revised as intermediate analyses reveal uncertainty or disagreement. 
This insight directly motivates \textbf{DDIAgents}, which models expert instantiation and knowledge routing as per-instance and iterative decisions, enabling targeted, interpretable and robust DDI prediction.

\begin{figure*}[t]
	\centering
	\includegraphics[width=\textwidth]{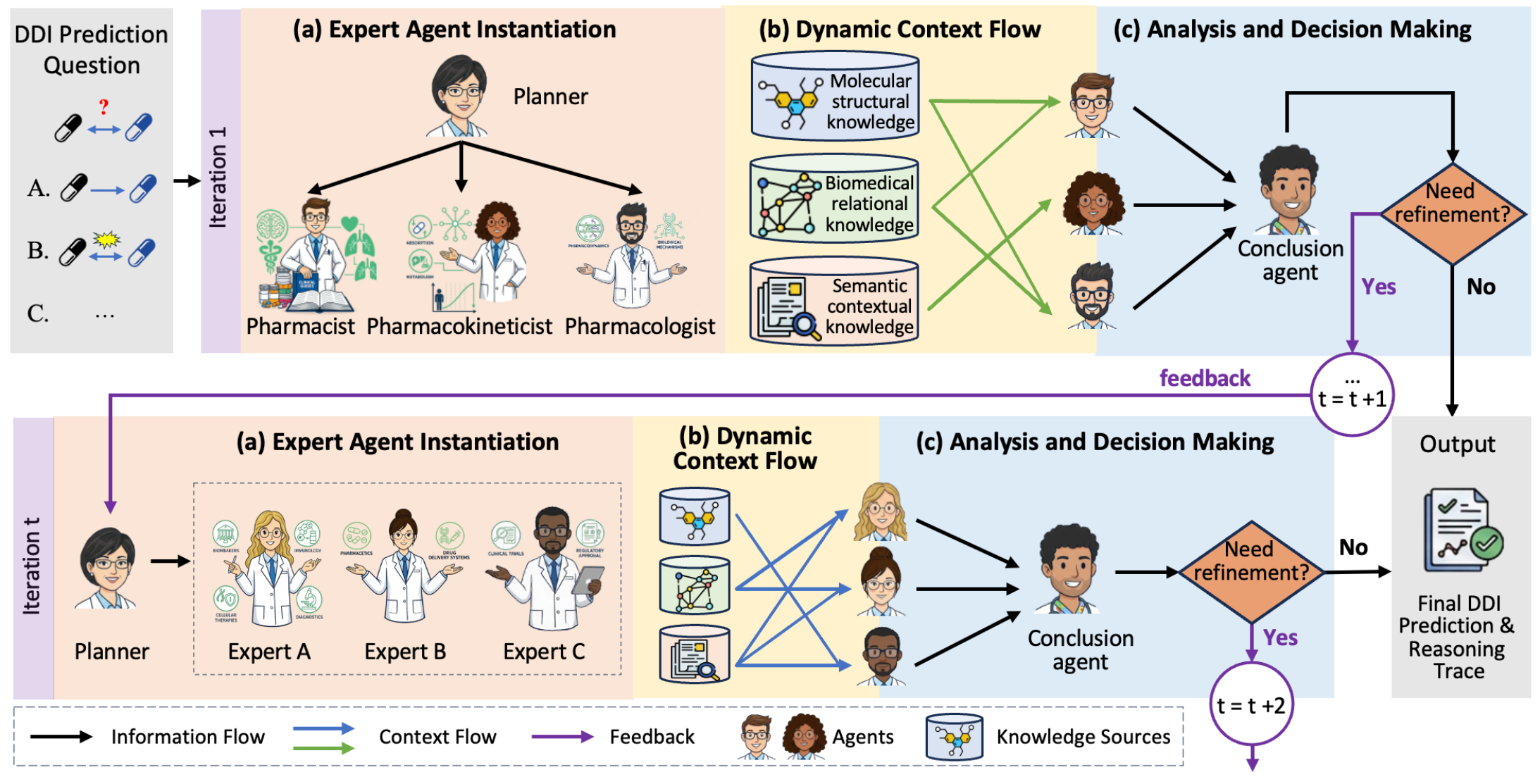}
	
	\vspace{-10px}
	\caption{The overall framework of the proposed DDIAgents, which iteratively conducts three stages: (a) Expert agent instantiation, (b) Dynamic context flow, (c) Analysis and decision making. }
	\label{fig:intro}
	\vspace{-10px}
\end{figure*}

\section{Methodology}


\subsection{Problem Setup}

Let $\mathcal{D}$ be the set of all drugs and $\mathcal{R}$ be the set of possible interaction types. 
In the DDI prediction problem, effective utilization of heterogeneous knowledge sources is critical.
We define a collection of knowledge sources as $\mathcal{S}$ that are relevant to the reasoning process in DDI prediction. Formally,
the DDI prediction problem is to learn a predictor $p$ that maps a pair of drugs and the integrated knowledge sources to a DDI type. 
For any drug pair $(u, v) \in \mathcal{D} \times \mathcal{D}$ and knowledge sources $\mathcal{S}$, the predicted DDI type $r \in \mathcal{R}$ is given by:
\begin{equation}
	r = p(u, v; \mathcal{S})
\end{equation}

For agent-based methods, we implement a filtering stage to handle the context overflow caused by an exhaustive list of DDI candidates. 
A traditional model~\citep{su2024dual, xu2024ddi} is trained to filter a predefined number of the most probable DDI candidates. 
The DDI prediction problem and these candidates are then formatted into a multiple-choice question $q$, serving as the input for agent-based methods.

\subsection{Overall Framework}
\label{sec:frame}

The framework of DDIAgents~(as shown in Figure~\ref{fig:intro}) incorporates the following three stages: 
\textbf{(a) Expert agent instantiation:} given a DDI query, a planner agent instantiates a tailored set of expert agents with distinct expertise. 
This enables adaptive expertise composition, avoiding rigid agent configurations.
\textbf{(b) Dynamic context flow:} the planner agent assigns each expert a mechanism-relevant subset of knowledge sources. This mitigates information overload and supports targeted and specialized reasoning.
\textbf{(c) Analysis and decision making:} expert agents conduct specialized analyses, and a conclusion agent synthesizes their outputs to either deliver a final prediction or request another analysis iteration.

If the conclusion agent detects unresolved uncertainty or conflicting evidence, its feedback serves as an update signal for the next iteration, guiding expert agent instantiation and dynamic context flow to avoid repetition of the same reasoning pattern.

\subsection{Expert Agent Instantiation}
\label{sec:recruit}

Selecting appropriate experts is essential for reliable and mechanism-aware DDI interpretation.
In the first iteration~($t=1$), we instantiate three core experts to cover complementary perspectives on clinical risk, pharmacokinetics, and biological mechanism:
\begin{itemize}[leftmargin=*]
\item Pharmacist~\citep{weideman1999pharmacist}: Focuses on practical, patient-oriented risks and outcomes of drugs in clinical usage. 
\item Pharmacokineticist~\citep{yu2019mechanisms}: Analyzes absorption, distribution, metabolism and excretion to identify exposure-altering interactions.
\item Pharmacologist~\citep{niu2019pharmacodynamic}: Reasons about pharmacodynamic properties and mechanism-level biological effects.
\end{itemize}

For subsequent iterations~($t > 1$), the framework moves beyond this fixed configuration. 
Instead of reusing the same roles, the planner analyzes the original query together with the conclusion agent's feedback from iteration $t-1$ to instantiate a new set of experts.
Concretely, the planner agent identifies unresolved gaps or conflicts in the last iteration, and then specifies expertise profiles via targeted role instructions (see Appendix~\ref{app:prompt}).
This adaptive instantiation allows DDIAgents to revise its reasoning trajectory to match the case-specific difficulty, rather than applying a fixed analytic pattern across all drug pairs.

\begin{figure*}[t]
	\centering
	\includegraphics[width=1.0\textwidth]{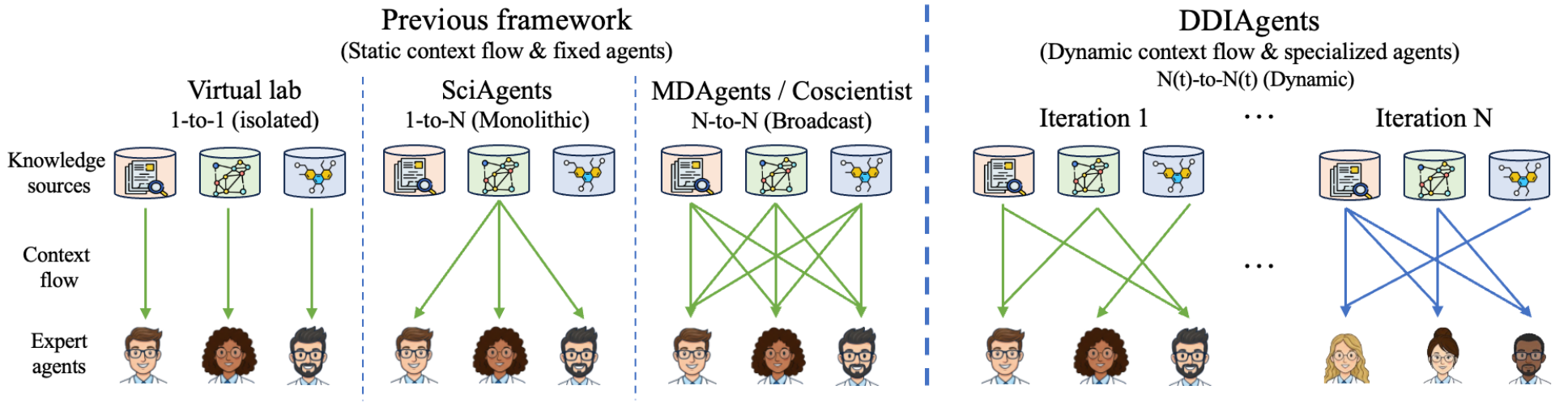}
	\vspace{-15px}
	\caption{Comparison between DDIAgents framework and previous multi-agent methods. }
	\vspace{-10px}
	\label{fig:method_comparison}
\end{figure*}

\subsection{Dynamic Context Flow}

To support specialized and rigorous expert analyses, we introduce a dynamic context flow that routes knowledge sources to experts in an expertise-aware and mechanism-dependent manner.
Concretely, we organize domain knowledge into a set of heterogeneous knowledge sources, each capturing a complementary view:
\begin{itemize}[leftmargin=*]
	\item \textbf{Molecular Structural Knowledge}: We incorporate the intrinsic chemical topology of drugs. Specifically, we utilize SMILES strings to represent the atom-bond connectivity and functional groups of drug molecules, which are fundamental to understanding the steric and electronic basis of interactions.

	\item \textbf{Biomedical Relational Knowledge}: We leverage the systemic connectivity of drugs within the biological network. This encompasses known DDI training data, drug-target associations and drug side-effect profiles. This view places the drug within the broader context of the biomedical system.
	
	\item \textbf{Semantic Contextual Knowledge}: We utilize natural language descriptions to capture high-level pharmacological concepts. This includes concise summaries of drug mechanisms and interaction types generated by LLMs~\citep{achiam2023gpt}, providing a semantic layer that complements raw structural data.
\end{itemize}

Given these knowledge sources, the dynamic context flow controls how they are assigned to expert agents. 
Specifically, the planner agent assigns each expert agent a subset of knowledge sources that aligns with the agent's expertise and the current reasoning demands.
As a result, agents may receive different combinations of evidence within the same iteration.
For instance, a chemistry-oriented expert may be assigned primarily structural evidence, while a biology-oriented expert may focus on relational evidence, and a clinical expert may rely more on semantic descriptions and safety-oriented relational cues. 
This allocation is updated across iterations using the intermediate feedback from the conclusion agent, enabling the system to shift attention as the hypothesized mechanism sharpens or new uncertainties emerge. 

By turning knowledge allocation into a targeted, expertise-aware pipeline, dynamic context flow mitigates two common failure modes in DDI reasoning. 
Firstly, it reduces information dilution~\citep{shi2023large}, where mechanism-critical evidence is buried in irrelevant context. 
Secondly, it alleviates cross-domain distraction~\citep{shen2024mome}, where experts are forced to process outside their functional role. 
This design prepares for the next stage of the framework: since each expert receives a role-aligned subset of knowledge sources, the subsequent analysis step can construct a tailored evidence context per expert and elicit deeper, more focused reasoning.

\subsection{Analysis and Decision Making}

At each iteration $t$, before expert agents begin reasoning, the system constructs a tailored knowledge context for them. 
This context is formed by retrieving and filtering the most pertinent evidence from the knowledge sources assigned in the current iteration
(see Appendix~\ref{app:retrieve}). 
The goal is to provide a focused but sufficient knowledge input, which is rich enough to support a mechanism-grounded analysis, while avoiding unnecessary information that can obscure critical cues.

Given the query and its curated context, each expert produces a DDI prediction and a structured rationale grounded in the provided evidence, reflecting its domain perspective. 
These expert reports are then passed to a conclusion agent, which synthesizes the set of analyses into a single decision. 
Rather than treating expert outputs as independent opinions, the conclusion agent aggregates them into a coherent explanation that resolves cross-expert dependencies. 

The conclusion agent has two types of possible outputs. 
If the evidence and expert analyses are sufficient, it finalizes the DDI prediction along with a consolidated reasoning trace. 
Otherwise, it produces iteration feedback that explicitly identifies what remains uncertain and recommends how the next iteration should adjust—both in terms of which new experts to instantiate and which knowledge sources to prioritize. 
This feedback closes the loop, ensuring that subsequent iterations are guided by concrete suggestions rather than repeating the same reasoning pattern.

\subsection{Comparison with Existing Works}

Table~\ref{tab:knowledge_comp} reveals a key limitation of existing DDI prediction methods: they typically pre-commit to a narrow knowledge source scope.
Feature-based methods mainly depend on molecular structure, graph-based methods incorporate biomedical networks but usually exclude unstructured descriptions, and LLM-based approaches usually depend on text alone. 
As a result, when the decisive cue for a drug pair lies outside the model's preferred modality, predictions can fail due to missing evidence.
In contrast, DDIAgents simultaneously considers all of these knowledge sources, and routes them to experts in a mechanism-aware, instance-adaptive manner.
Figure~\ref{fig:method_comparison} further contrasts DDIAgents with representative multi-agent frameworks for scientific tasks.
Unlike existing systems that rely on static execution, DDIAgents adopts an adaptive and iterative reasoning framework. 
Rather than maintaining fixed agent roles, DDIAgents leverages specialized agent instantiation to generate specialized experts tailored to specific sub-problems as the task evolves.
The most significant distinction lies in the context flow, with different systems applying different mapping patterns: 
\begin{itemize}[leftmargin=*]
	\item 1-to-1 (Isolated): Virtual Lab~\citep{swanson2025virtual} assigns specific, pre-defined knowledge sources to specific agents in a fixed, parallel mapping.
	\item 1-to-N (Monolithic): SciAgents~\citep{ghafarollahi2025sciagents} relies on a single, monolithic knowledge graph distributed across multiple agents.
	\item N-to-N (Broadcast): Coscientist~\citep{boiko2023autonomous} and MDAgents~\citep{kim2024mdagents} distribute all available information to every agent uniformly. 
	\item N(t)-to-N(t) (Dynamic): DDIAgents treats the relation between knowledge sources and agents as variables that evolve over time. It adaptively maps pertinent, domain-specific data to the relevant experts based on the current reasoning state. 
\end{itemize}

\begin{table}[t]
	\centering
	\setlength{\tabcolsep}{2pt}
	\caption{Knowledge source utilization comparison.}
	\vspace{-10px}
	\begin{tabular}{l|ccc}
		\hline
		\diagbox[width=11em]{\textbf{Method}}{\textbf{Knowledge}} & 
		\specialcell{\textbf{Molecular}\\\textbf{Structure}} & 
		\specialcell{\textbf{Biomedical}\\\textbf{Network}} & 
		\specialcell{\textbf{Text}\\\textbf{Description}} \\
		\hline
		Feature-based~\citep{ryu2018deep, liu2022predict} & \cmark & {-} & {-} \\
		Graph-based~\citep{nyamabo2021ssi, zitnik2018modeling} & \cmark & \cmark & {-} \\
		LLM-based~\citep{zhu2023learning, xu2024ddi} & {-} & {-} & \cmark \\
		\hline
		\textbf{DDIAgents (Ours)} & \textbf{\cmark} & \textbf{\cmark} & \textbf{\cmark} \\ \hline
	\end{tabular}
	\label{tab:knowledge_comp}
\end{table}

\section{Experiments}

\begin{table*}[t]
  \centering
  \setlength{\tabcolsep}{3.5pt}
  \caption{General performance comparison between different DDI prediction methods. Results are reported as mean$\pm$std over 5 independent runs. The best and second-best results are highlighted in bold and underline, respectively. 
  }
  
  \vspace{-10px}
  \begin{tabular}{c|l|cccc|cccc}
    \hline
    &&\multicolumn{4}{c|}{\textbf{DrugBank}}&\multicolumn{4}{c}{\textbf{TWOSIDES}} \\
    \textbf{Type} & \textbf{Method} & \multicolumn{2}{c}{S1} & \multicolumn{2}{c|}{S2} & \multicolumn{2}{c}{S1}  & \multicolumn{2}{c}{S2} \\
    &&Acc&F1&Acc&F1&Hit@5&NDCG@5&Hit@5&NDCG@5 \\\hline
    Feature-based & MLP & 34.25$\pm$1.15 & 11.00$\pm$0.81 & 20.93$\pm$1.29 & 2.80$\pm$0.41 & 40.08$\pm$2.19 & 14.29$\pm$0.72 & 26.79$\pm$0.35 & 7.24$\pm$0.47 \\ \hline
    \multirow{4}{*}{Graph-based} & Decagon & 20.07$\pm$1.59 & 3.49$\pm$0.54 & 6.87$\pm$0.22 & 1.96$\pm$0.11 & 32.75$\pm$1.13 & 10.71$\pm$0.57 & 24.22$\pm$0.65 & 6.59$\pm$0.29 \\
    & EmerGNN & 29.57$\pm$1.43 & 13.32$\pm$0.37 & 15.33$\pm$0.49 & 3.54$\pm$0.07 & 35.84$\pm$1.75 & 11.07$\pm$0.68 & 28.35$\pm$1.34 & 8.56$\pm$0.72 \\
    & MSTE & 31.60$\pm$0.37 & 12.87$\pm$0.12 & 12.73$\pm$0.14 & 1.80$\pm$0.07 & 30.94$\pm$0.62 & 10.69$\pm$0.14 & 25.00$\pm$1.13 & 6.67$\pm$0.45 \\
    & TIGER & 45.98$\pm$2.35 & 26.78$\pm$1.33 & 20.64$\pm$1.44 & 3.32$\pm$0.37 & 42.24$\pm$0.38 & 15.00$\pm$0.21 & 25.26$\pm$0.17 & 6.65$\pm$0.09 \\ \hline
    \multirow{5}{*}{LLM-based} & Single LLM & 8.71$\pm$0.37 & 4.10$\pm$0.32 & 7.30$\pm$0.29 & 3.94$\pm$0.23 & 27.30$\pm$0.31 & 7.91$\pm$0.17 & 23.78$\pm$0.74 & 6.58$\pm$0.36 \\ 
    & K-Path & 35.34$\pm$1.08 & 21.41$\pm$0.59 & 19.48$\pm$0.48 & 3.71$\pm$0.19 & 31.15$\pm$0.43 & 9.57$\pm$0.18 & 24.01$\pm$0.38 & 5.82$\pm$0.07 \\ 
    & CBR-DDI & 41.38$\pm$1.68 & 31.47$\pm$0.76 & 23.42$\pm$0.41 & 4.48$\pm$0.47 & 35.07$\pm$1.51 & 13.09$\pm$0.47 & 32.77$\pm$0.34 & 10.31$\pm$0.18 \\ 
    & TextDDI & 49.53$\pm$2.14 & 31.72$\pm$1.27 & 28.75$\pm$0.91 & 8.26$\pm$0.37 & 42.04$\pm$1.86 & 15.38$\pm$0.79 & 29.35$\pm$1.18 & 9.77$\pm$0.75 \\ 
    & DDI-GPT & 50.68$\pm$0.79 & 34.71$\pm$1.07 & 33.47$\pm$1.21 & 15.01$\pm$0.61 & 43.90$\pm$2.61 & 17.31$\pm$1.44 & 33.51$\pm$1.34 & 10.24$\pm$0.68 \\ 
    \hline
    \multirow{5}{*}{Agent-based} & Reflexion & 47.31$\pm$2.34 & 35.11$\pm$2.45 & 31.74$\pm$1.76 & 14.07$\pm$1.22 & 41.47$\pm$1.92 & 17.04$\pm$1.21 & 30.71$\pm$1.44 & 9.55$\pm$0.51 \\ 
    & Debate & 52.18$\pm$1.46 & 40.14$\pm$1.17 & 35.03$\pm$1.64 & 15.84$\pm$0.78 & 50.61$\pm$2.07 & 19.03$\pm$0.74 & 38.66$\pm$1.41 & 12.78$\pm$0.62 \\ 
    & AgentVerse & 52.34$\pm$2.08 & \underline{41.17$\pm$1.21} & 36.48$\pm$0.56 & 14.01$\pm$0.38 & 49.33$\pm$1.46 & 19.44$\pm$0.59 & \underline{40.38$\pm$0.76} & \underline{13.14$\pm$0.24} \\ 
    & MDAgents & \underline{54.69$\pm$0.74} & 39.52$\pm$0.54 & \underline{36.29$\pm$1.17} & \underline{15.97$\pm$0.42} & \underline{51.42$\pm$2.14} & \underline{20.07$\pm$1.36} & 40.09$\pm$1.61 & 12.64$\pm$0.75 \\ 
    \cline{2-10}
    & DDIAgents & \textbf{55.75$\pm$0.74} & \textbf{45.52$\pm$0.89} & \textbf{38.21$\pm$1.03} & \textbf{18.17$\pm$0.51} & \textbf{53.94$\pm$1.24} & \textbf{21.21$\pm$0.48} & \textbf{42.26$\pm$0.92} & \textbf{14.08$\pm$0.15} \\ \hline
  \end{tabular}
  \label{tab:general_performance}
\end{table*}

\subsection{Experimental Setup}
\label{sec:exp_setup}

\textbf{Datasets. } Following~\citep{yu2021sumgnn, zhang2023emerging}, we conduct experiments on two widely used public DDI prediction datasets: DrugBank~\citep{wishart2018drugbank}, a multiclass DDI prediction dataset where each drug pair is associated with one of 86 possible interaction types; TWOSIDES~\citep{tatonetti2012data}, a multilabel DDI prediction dataset that records 209 possible side effects.

\noindent \textbf{Experimental Setting. }
To incorporate realistic factors in evaluation, we partition the drug set $\mathcal{D}$ into known drugs $\mathcal{D}_\textnormal{known}$ and new drugs $\mathcal{D}_\textnormal{new}$ based on approval time.
We consider three prediction tasks: S0 (known--known), S1 (known--new), and S2 (new--new).
In this work, we focus on the more realistic and challenging S1/S2 tasks, and report S0 results in Appendix~\ref{sec:s0}.

\noindent \textbf{Evaluation Metrics.} For the DrugBank dataset that features a single interaction type per drug pair, we employed Accuracy and F1 Score. 
The TWOSIDES dataset often includes multiple interaction types per drug pair; thus, we framed the analysis as a recommendation task, using Hit@5 and NDCG@5 for evaluation. 
This evaluation method differs from most existing works that use TWOSIDES for binary classification~(results shown in Appendix~\ref{app:two}).
We argue that binary classification tasks are often redundant and include uninformative negative samples. 
Our proposed ranking metrics accurately reflect realistic scenarios where a drug pair may exhibit multiple interactions, and the objective is to identify the most probable risks.

\noindent \textbf{Implementation Details.} 
We use Qwen2.5 series~\citep{qwen2.5} as the backbone LLM for DDIAgents and other types of agent-based baseline methods in quantitative experiments. 
We set the maximum iteration round $T=3$ and agent number $m=3$. 
For DrugBank, each DDI question takes 12.83 seconds on average, and for TWOSIDES it takes 15.66 seconds—both within an acceptable range.
More implementation details are shown in Appendix~\ref{sec:imp_dtl}. 
Our code and data are available at \url{https://github.com/zgs0314/DDIAgents}.

\noindent \textbf{Baseline Methods.} In this work, we compare our proposed DDIAgents framework with four types of baseline methods: (1) \textit{Feature-based methods}: MLP~\citep{rogers2010extended}; (2) \textit{Graph-based methods}: MSTE~\citep{yao2022effective}, Decagon~\citep{zitnik2018modeling}, EmerGNN~\citep{zhang2023emerging} and TIGER~\citep{su2024dual}; (3) \textit{LLM-based methods}: Single LLM~\citep{hong2023metagpt}, TextDDI~\citep{zhu2023learning}, DDI-GPT~\citep{xu2024ddi}, CBR-DDI~\citep{liu2025case}, K-Path~\citep{abdullahi2025k}; and (4) \textit{Agent-based methods}: Reflexion~\citep{shinn2023reflexion}, Debate~\citep{du2023improving}, AgentVerse~\citep{chen2023agentverse}, MDAgents~\citep{kim2024mdagents}.




\subsection{General Performance Comparison}
Table~\ref{tab:general_performance} summarizes the overall comparison across S1 and S2 tasks on DrugBank and TWOSIDES.
We can see that our proposed DDIAgents framework consistently outperforms all baseline methods. 
Compared with the best-performing baseline, DDIAgents improves F1 by 13.77\% in S2 tasks on DrugBank, and increases NDCG@5 by 7.15\% in S2 tasks on TWOSIDES.
These gains validate the core design of DDIAgents: orchestrating multiple expert agents and routing specialized knowledge sources to each expert to enable complementary and targeted reasoning.

Among baselines, feature-based and graph-based models show comparatively weak performance. 
A key reason is that the evaluation protocol adopts an approval-time-based drug split to better reflect real-world deployment, which substantially increases task difficulty. 
Under this more realistic setting, methods that rely on molecular features or limited graph structure cannot effectively leverage heterogeneous evidence, and thus struggle to capture the diverse mechanisms underlying DDIs.
The performance of LLM-based methods also remains limited, primarily because they utilize only the textual descriptions, which fails to comprehensively model the multi-faceted evidence required for robust DDI prediction. 
In particular, the Single LLM baseline performs poorly, suggesting that memorization or knowledge leakage in current LLMs is not the primary driver of strong performance in this task. 

In contrast, agent-based methods generally achieve the best performance among different types of methods. 
This is because their strong capability in handling heterogeneous knowledge sources and the shared high-recall candidate retriever provides a reasonable starting point for the prediction~(its performance is shown in Appendix~\ref{app:trad}). 
Notably, methods that incorporate multiple expert agents (e.g., MDAgents and AgentVerse) consistently outperform single-agent methods (e.g., Reflexion). 
This comparison highlights the critical need for integrating specialized analyses from multiple complementary perspectives.

\begin{figure*}[t]
	\centering
	\includegraphics[width=0.95\textwidth]{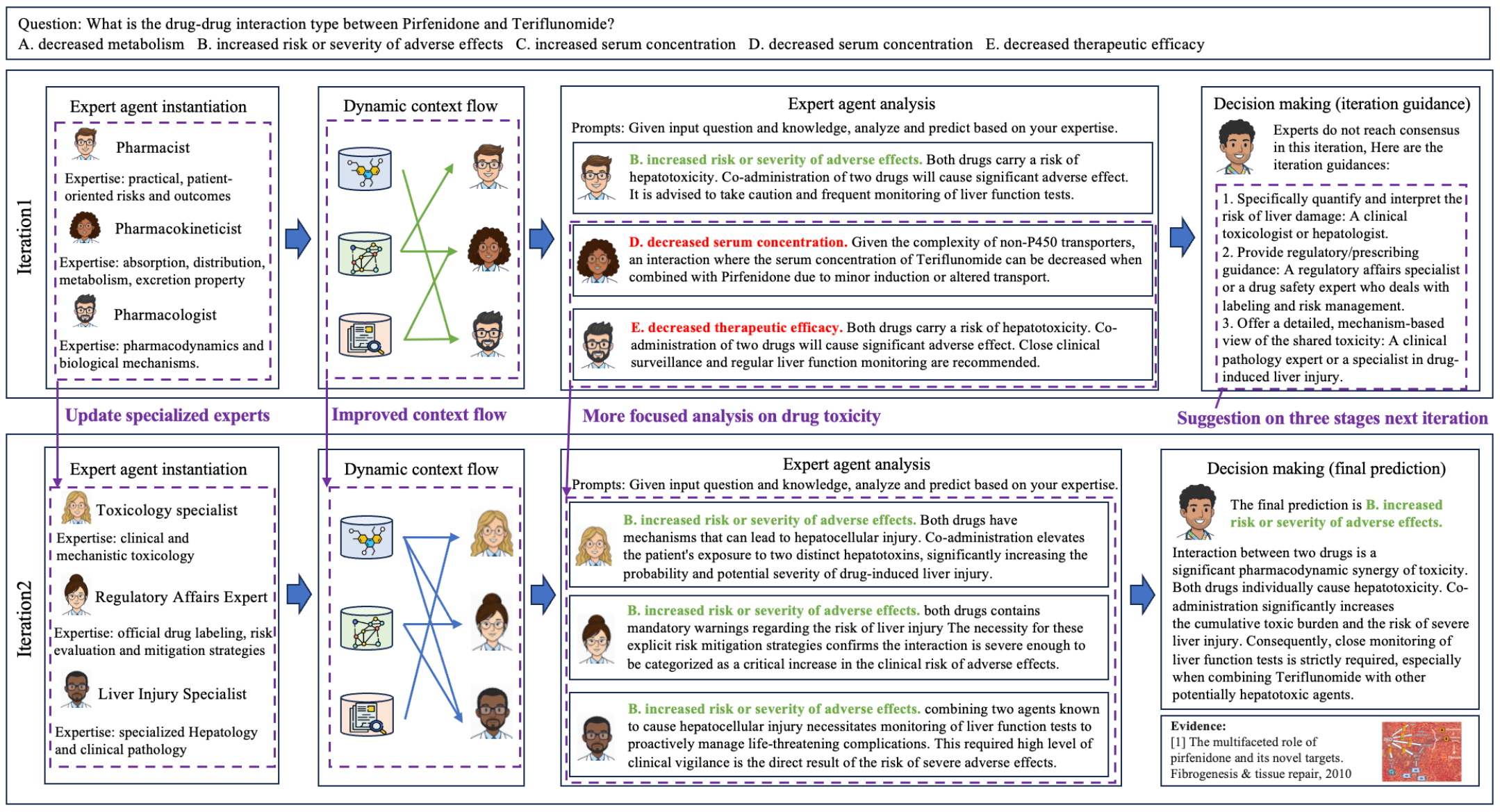}
  \vspace{-10px}
	\caption{A prediction case of DDIAgents. Bold red text indicates incorrect answers and bold green text indicates correct answers. }
	\label{fig:case}
\end{figure*}

\subsection{Ablation Study}
To validate the effectiveness of the proposed components in DDIAgents, we compare it with the following variants: 
(a) w/o DynExp, which utilizes three predefined expert agents across iterations; 
(b) w/o DynKnow, which provides all of the knowledge sources instead of using dynamic context flow; 
(c) w/o Iter, which only uses a single iteration to obtain the DDI prediction.

From results in Table~\ref{tab:ablation}, we can see that removing any proposed component results in a performance drop, validating the necessity of the designs in DDIAgents.
Specifically, the largest performance degradation is in the "w/o DynKnow" variant, highlighting the critical importance of the dynamic context flow. 
Furthermore, we observe that the iterative analysis component is more crucial for the TWOSIDES dataset. 
We hypothesize this is because the DDI prediction task on TWOSIDES requires selecting multiple DDI types from a set of candidates, leading to a higher frequency of inter-expert disagreement and, consequently, making the iterative refinement process essential for consensus and accurate prediction.

\begin{table}[h]
  \centering
  \setlength{\tabcolsep}{1.2pt}
  \caption{Comparison of different variants of DDIAgents}
  
  \vspace{-10px}
  
  \begin{tabular}{c|cccc|cccc}
    \hline
    &\multicolumn{4}{c|}{\textbf{DrugBank}}&\multicolumn{4}{c}{\textbf{TWOSIDES}} \\
     & \multicolumn{2}{c}{S1} & \multicolumn{2}{c|}{S2} & \multicolumn{2}{c}{S1}  & \multicolumn{2}{c}{S2} \\
    &Acc&F1&Acc&F1&Hit&NDCG&Hit&NDCG \\\hline
    DDIAgents & 55.75 & 45.52 & 38.21 & 18.17 & 53.94 & 21.21 & 42.26 & 14.08 \\
    \hline
    w/o DynExp & 54.08 & 44.39 & 36.70 & 16.66 & 51.73 & 20.37 & 41.01 & 13.52 \\
    w/o DynKnow & 52.23 & 40.19 & 35.57 & 16.01 & 50.34 & 19.52 & 38.71 & 12.07 \\
    w/o Iter & 53.68 & 42.15 & 36.67 & 16.71 & 50.72 & 19.44 & 39.56 & 12.88 \\
    \hline
  \end{tabular}
  \label{tab:ablation}
\end{table}


\subsection{Case Study}
\label{exp:case}

Figure~\ref{fig:case} shows an exemplar case of DDIAgents on a DDI prediction problem on DrugBank dataset using large-scale frontier model~(GPT-4o). 
The case study details a two-iteration process to predict the DDI type between Pirfenidone and Teriflunomide, ultimately identifying the correct answer.

In the first iteration, the DDIAgents framework utilizes predefined pharmacist, pharmacokineticist, and pharmacologist as expert agents to analyze the question, but they failed to reach a consensus. 
In this case, the conclusion agent identifies conflicting evidence and generates targeted iteration guidance.
Its feedback calls for instantiating toxicology and regulatory experts, identifies regulatory labeling as a key source for dynamic context flow, and prioritizes liver toxicity as the main focus for subsequent expert analysis.

Acting on this directive, expert agent instantiation in the second iteration gathers toxicology specialist, regulatory affairs expert, and liver injury specialist as updated experts better suited to the problem.
The dynamic context flow then prioritizes official drug labeling and mechanistic toxicology evidence to fill the identified gaps.
With these refined inputs, expert analysis converges on a unified understanding of the mechanisms underlying hepatocellular injury.
Finally, the decision-making stage confirms the prediction of ``increased risk or severity of adverse effects'', 
supported by an explanation of synergistic toxic burden and the clinical need to monitor liver function tests.
To verify the reliability, we performed a secondary search of relevant literature, 
confirming that output analysis is well-supported by established clinical evidence~\citep{macias2010multifaceted}. 
More DDI prediction cases using Qwen 2.5 7B are shown in Appendix~\ref{app:case}. 

\subsection{Performance across DDI Type Frequencies}
In this section, we evaluate the robustness of DDIAgents framework across various DDI types using the DrugBank dataset.
Representative methods from different categories are selected for comparison, including MLP, TIGER, DDI-GPT, Reflexion, MDAgents and DDIAgents. 
Given the inherent class imbalance, we employ a frequency-based ranking to categorize DDI types into five quintile bins based on frequency. 
The ``0\%--20\%'' bin contains the most frequent DDI types, while the ``80\%--100\%'' bin represents the long-tail DDI types.

\begin{figure}[h]
	\centering
	\includegraphics[width=0.45\textwidth]{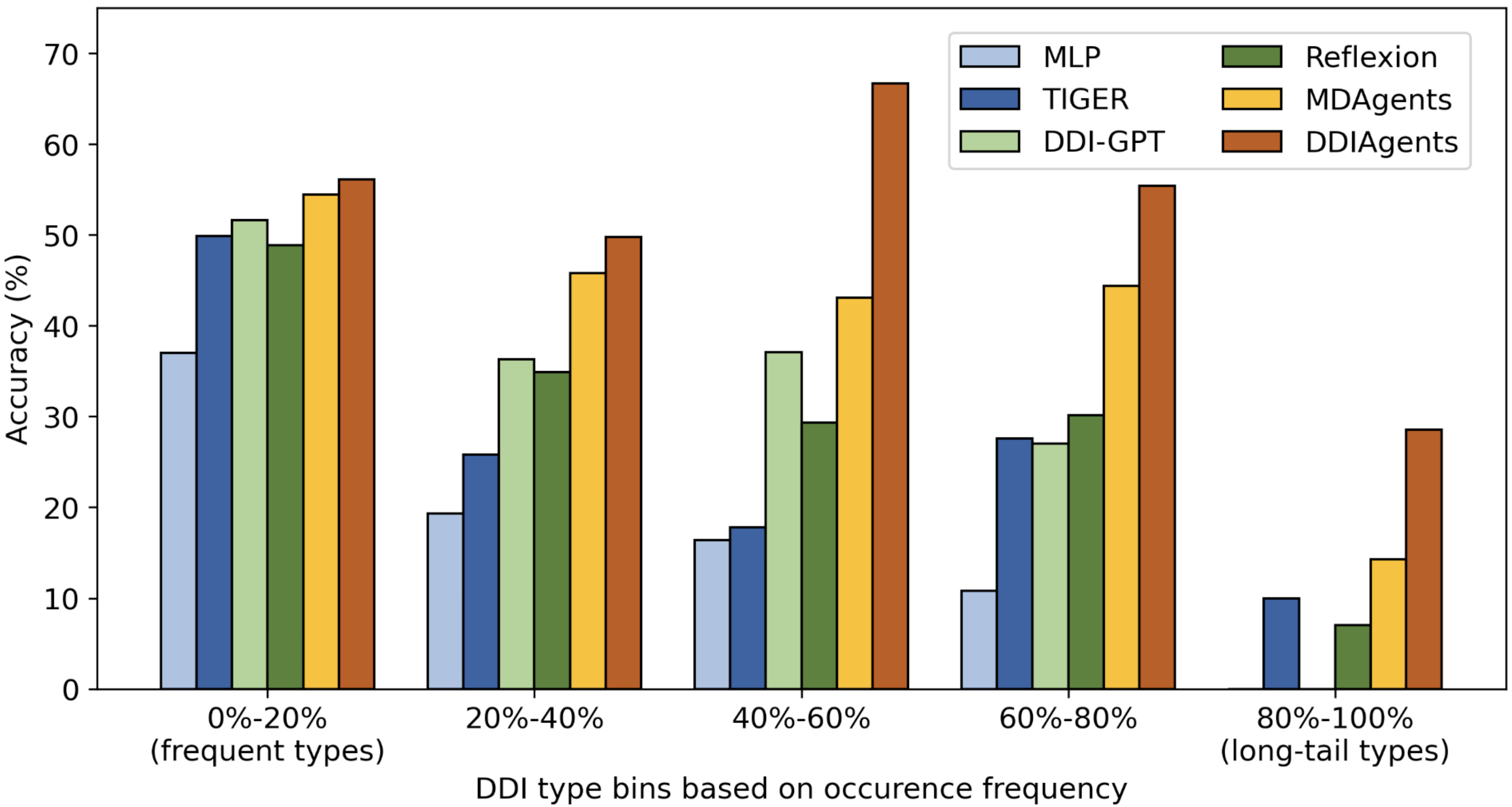}
  \vspace{-10px}
	\caption{DDI prediction performance across DDI type frequency bins on DrugBank S1 tasks. }
  \vspace{-5px}
	\label{fig:ddi_type}
\end{figure}

Figure~\ref{fig:ddi_type} illustrates the average prediction accuracy across DDI type bins on S1 tasks.
The results demonstrate that DDIAgents consistently outperforms all baseline methods across every frequency interval. Notably, the performance advantage of DDIAgents becomes increasingly significant on long-tail DDI types.
This trend suggests that while traditional models depend on high-frequency patterns, DDIAgents better handles rare, data-sparse interactions through collaborative agents and dynamic context flow.

\subsection{Performance across Different LLM Models}

\begin{figure}[ht]
    \centering
    \begin{subfigure}{0.24\textwidth}
        \centering
        \includegraphics[width=\linewidth]{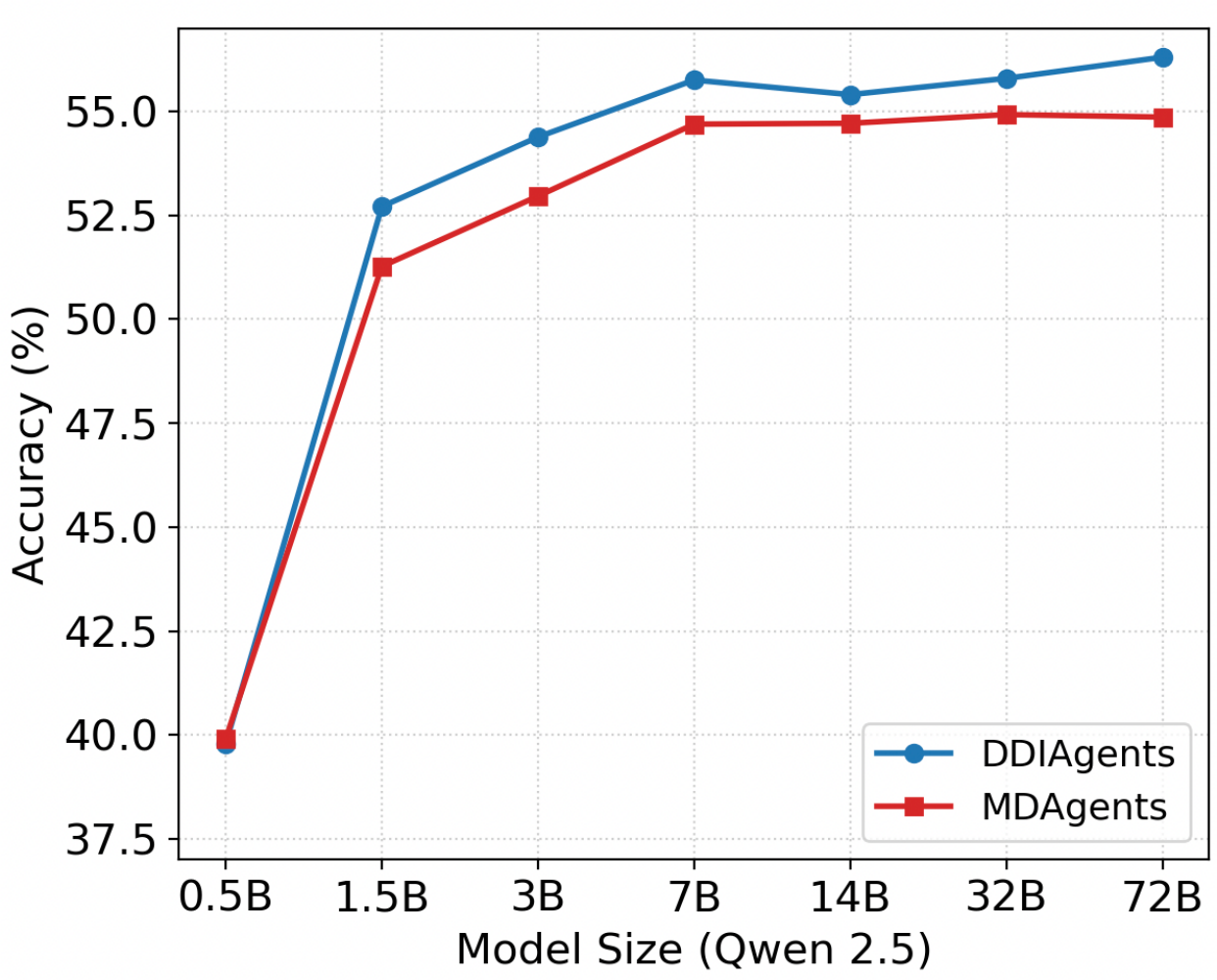}
        \caption{S1 tasks}
        \label{fig:acc_comparison}
    \end{subfigure}
    \hfill 
    \begin{subfigure}{0.23\textwidth}
        \centering
        \includegraphics[width=\linewidth]{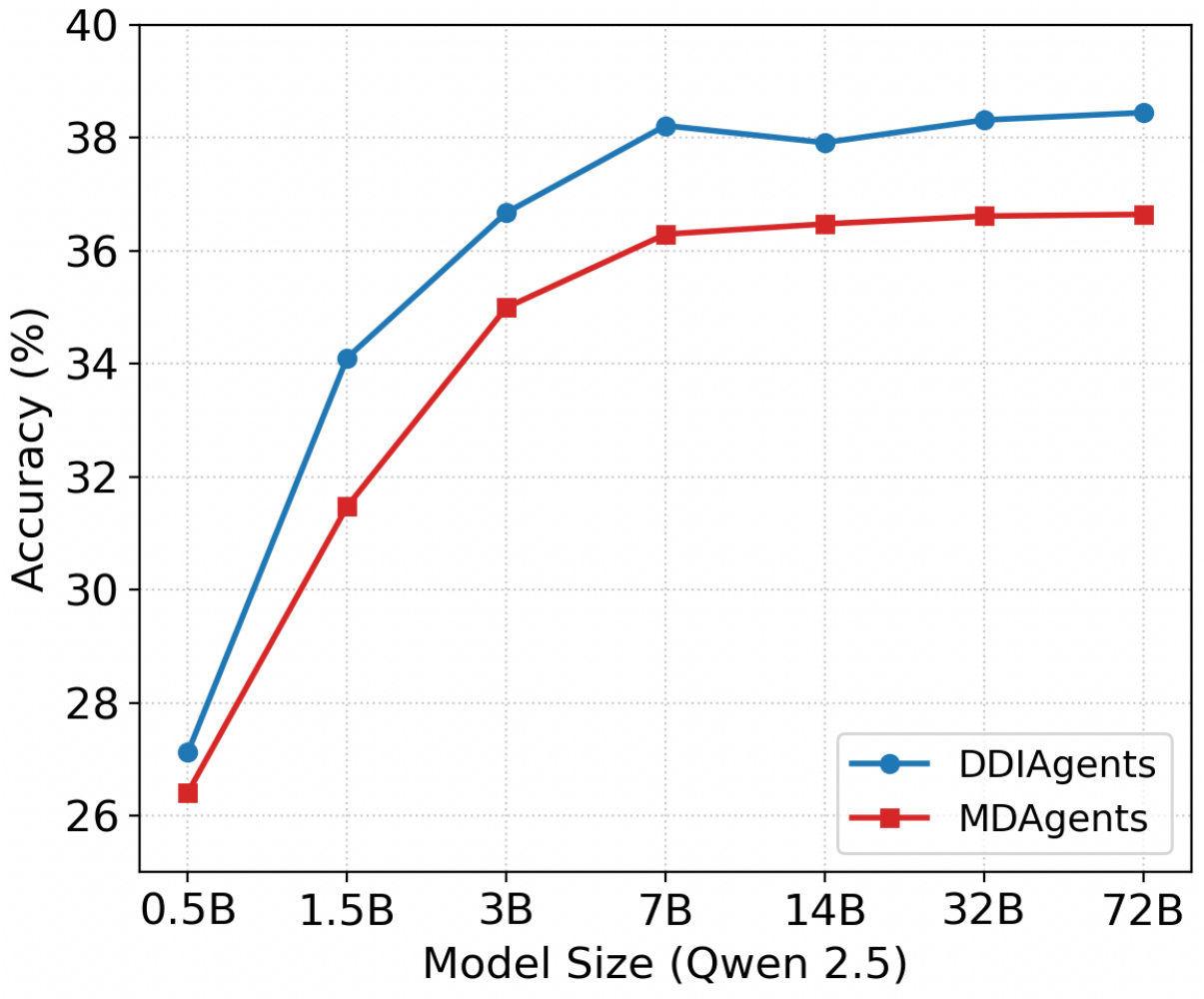}
        \caption{S2 tasks}
        \label{fig:f1_comparison}
    \end{subfigure}
    
    \vspace{-10px}
    \caption{The performance of DDIAgents and MDAgents with different LLM model sizes on the DrugBank dataset. }
    \vspace{-5px}
    \label{fig:llm_size}
\end{figure}

To rigorously validate the generalizability and inherent robustness, we conducted an extensive evaluation on DDIAgents using various LLMs as agents and compared them with the strongest baseline, MDAgents.
Our initial set of experiments focused on assessing the influence of model scale within a unified family, utilizing the Qwen 2.5 series with sizes ranging from 0.5B to 72B parameters. 
The results in Figure~\ref{fig:llm_size} show that DDIAgents outperforms MDAgents at most tested model sizes, and that both frameworks improve steadily as model scale increases.
Notably, the smaller models (0.5B and 1.5B) were unable to achieve satisfactory prediction performance for both methods, underscoring the requirement for a sufficiently capable LLM base to execute the complex tasks within the multi-agent system. 
Furthermore, DDIAgents achieves higher performance as model size increased, which suggests a greater capacity to stimulate the emergent capabilities of large-scale LLMs.

To evaluate the cross-architecture generalizability, we compared the performance of DDIAgents and MDAgents across LLMs of comparable scales, including Qwen 2.5 7B, Llama 3.1 8B \citep{dubey2024llama}, Gemma 2 9B \citep{team2024gemma} and ChatGLM 3 6B \citep{du2022glm}.
The results in Table~\ref{tab:llm_type} indicate that DDIAgents consistently outperforms MDAgents, demonstrating superior robustness and reasoning capabilities.
Notably, while both frameworks perform well with Qwen 2.5, Llama 3.1 and Gemma 2, DDIAgents maintains a substantial lead in F1 scores, which indicates its enhanced capacity for analyzing long-tail DDI types.
The relatively low performance of ChatGLM 3 may stem from its training that is less optimized for complex biomedical reasoning.

\begin{table}[h]
  \centering
  \setlength{\tabcolsep}{4pt}
  \caption{The performance of DDIAgents and MDAgents using LLM models with similar sizes on DrugBank dataset. }
  
  \vspace{-10px}
  \begin{tabular}{c|c|cccc}
    \hline
    & &\multicolumn{4}{c}{\textbf{DrugBank}} \\
    LLM models & Method & \multicolumn{2}{c}{S1} & \multicolumn{2}{c}{S2}  \\
    &&Acc&F1&Acc&F1 \\\hline
    \multirow{2}{*}{Qwen 2.5 7B} & DDIAgents & 55.75 & 45.52 & 38.21 & 18.17 \\ 
    &MDAgents&54.69 & 39.52 & 36.29 & 15.97\\\hline
    \multirow{2}{*}{Llama 3.1 8B}& DDIAgents & 55.63 & 46.56 & 37.21 & 17.77 \\ 
    &MDAgents&53.97&40.72&36.32&16.24\\\hline
    \multirow{2}{*}{Gemma 2 9B}& DDIAgents & 55.30 & 44.19 & 38.34 & 18.39 \\ 
    &MDAgents&54.36&40.41&35.72&16.04 \\\hline
    \multirow{2}{*}{ChatGLM 3 6B}& DDIAgents & 53.71 & 42.60 & 35.98 & 16.44 \\ 
    &MDAgents&52.84&38.11&34.78&15.19\\\hline
  \end{tabular}
  \label{tab:llm_type}
  \vspace{-5px}
\end{table}

\subsection{Comparison on Generated Contents}
\label{exp:interpertability}

To evaluate the quality of the generated explanations, we conducted a comparative analysis between DDIAgents and MDAgents. 
We employed a state-of-the-art LLM (GPT-4o) as a judge to evaluate their responses on S2 tasks across four domain-specific dimensions~\citep{guo2024lightrag}:
\begin{itemize}[leftmargin=*]
\item Thoroughness: how comprehensively the system addresses the multi-faceted nature of DDIs; 
\item Diversity: the breadth of explored biological mechanisms;
\item Rationality: whether the answer provides reasonable analysis; 
\item Overall: a holistic judgment of which system provides a more useful and trustworthy reference for drug interaction analysis.
\end{itemize}

As shown in Table~\ref{tab:reasonable}, DDIAgents consistently outperforms MDAgents across all dimensions. 
The expert agent instantiation phase enhances pharmacological thoroughness and mechanistic diversity by integrating specialized domain expertise into the reasoning process. 
Simultaneously, the dynamic context flow and iterative feedback loops refine agent outputs, significantly boosting clinical rationality. 
By replacing static discussions with active collaboration, DDIAgents captures complex biochemical interactions more accurately, providing more reliable and comprehensive DDI explanations than generic multi-agent frameworks.

\begin{table}[h]
  \centering
  \setlength{\tabcolsep}{2.5pt}
  \caption{Win rates (\%) for generated contents of DDIAgents vs. MDAgents across four  dimensions on S2 tasks.}
  \vspace{-10px}
  \begin{tabular}{l|cc|cc}
    \hline
    &\multicolumn{2}{c|}{\textbf{DrugBank}}&\multicolumn{2}{c}{\textbf{TWOSIDES}} \\
    & DDIAgents & MDAgents & DDIAgents & MDAgents\\ \hline
    Thoroughness & 66.39\% & 33.61\% & 72.27\% & 27.73\% \\ 
    Diversity & 67.38\% & 32.62\% & 67.33\% & 32.67\% \\
    Rationality & 75.02\% & 24.98\% & 70.87\% & 29.13\% \\
    Overall & 69.08\% & 30.92\% & 71.34\% & 28.66\% \\\hline
  \end{tabular}
  \label{tab:reasonable}
  \vspace{-10px}
\end{table}

\section{Conclusion}

In this work, we presented DDIAgents, a mechanism-conditioned multi-agent framework for DDI prediction. 
Rather than treating DDI analysis as static feature fusion or historical case reuse, DDIAgents frames the task as knowledge orchestration: for each drug pair, dynamic context flow routes heterogeneous biomedical evidence to specialized agents according to mechanism-specific reasoning needs. 
Extensive experiments demonstrate that DDIAgents consistently outperforms state-of-the-art baselines. 
Furthermore, DDIAgents provides interpretable agent-level mechanistic insights into predicted interactions, offering a useful direction for adaptive and evidence-grounded AI4Science reasoning. 

Despite these advancements, the current implementation relies on a fixed agent collaboration scheme, which may restrict the system's adaptability. 
Future work will develop a more flexible collaboration scheme with multiple interaction patterns based on the specific context. 
Additionally, we plan to incorporate more critical knowledge sources to expand the dynamic context flow and further enhance predictive accuracy and robustness.

\section{Limitations and Ethical Considerations}

\textbf{Limitations.} The current DDIAgents adopts a fixed agent collaboration scheme with predefined interaction protocols. 
While this simplifies orchestration, it may reduce adaptability when different DDI queries would benefit from alternative coordination patterns. 
In addition, DDIAgents currently relies on a limited set of knowledge-source types in the dynamic context flow, which may omit clinically decisive evidence. 

\noindent \textbf{Ethical considerations.} Our experiments use publicly available drug resources rather than patient-identifiable records. 
Nevertheless, any clinical deployment should ensure proper consent, governance and strong security for data access and storage. 
DDIAgents is intended for decision support but not final decisions. The outputs should be verified with reliable evidence to mitigate harms from hallucinations or misuse.

\section*{GenAI Disclosure}
This work uses Generative AI. 
The LLM-based baselines and agent-based methods rely on large language models to generate predictions and explanations for DDI queries. 
The case studies include representative model-generated outputs to illustrate the agent workflows. 
All such content is produced at inference time for research evaluation and demonstration, and may contain inaccuracies as discussed in the Limitations and Ethical Considerations section.

\section*{Acknowledgment}

Q. Yao is supported
Beijing Natural Science Foundation (No.4242039)
and
Beijing Science and Technology Program (No.Z251100008125003).

\bibliographystyle{ACM-Reference-Format}
\bibliography{reference}

\appendix





\clearpage


\section{Supplementary Information}
\label{sec:imp_dtl}

\subsection{Description of DDI Examples in Motivation Part}

In Table~\ref{tab:limit}, we provide textual descriptions of diverse DDI mechanisms illustrated in Figure~\ref{fig:limit}. 
Here we provide more comprehensive explanations on DDI mechanisms and details of reference knowledge used in exemplar cases. 

\begin{table}[h]
  \centering
  \small
  \caption{Three DDI examples with different mechanisms.}
  \begin{tabular}{p{8cm}}
    \hline
    \textbf{DDI examples}\\
    \hline
    \textbf{Drug 1:} Roxatidine acetate; \textbf{Drug 2:} Risedronic acid; \\
    \textbf{Interaction:} Increased serum concentration and adverse side effect. \\
    \textbf{Mechanism:} Roxatidine acetate raises the gastric $\text{pH}$, which unexpectedly increases the absorption of Risedronic acid, leading to a higher serum concentration and a greater risk of adverse effects. \\
    \textbf{Reference knowledge:} Drug usage instruction (the administration instructions for Risedronic acid) \\ \hline
    \textbf{Drug 1:} Esomeprazole; \textbf{Drug 2:} Clopidogrel; \\
    \textbf{Interaction:} Decreased antiplatelet activity. \\
    \textbf{Mechanism:} Esomeprazole, by inhibiting the liver enzyme $\text{CYP2C19}$, prevents the necessary metabolic conversion of the inactive Clopidogrel, consequently decreasing its antiplatelet activity. \\
    \textbf{Reference knowledge:} Biomedical networks (the drug-target interaction between esomeprazole and enzyme $\text{CYP2C19}$) \\ \hline
    \textbf{Drug 1:} Erythrityl tetranitrate; \textbf{Drug 2:} Sildenafil; \\
    \textbf{Interaction:} Increased vasodilatory activity. \\
    \textbf{Mechanism:} Erythrityl tetranitrate increases the concentration of the vasodilator $\text{cGMP}$ by donating $\text{NO}$, while Sildenafil prevents its breakdown by inhibiting $\text{PDE5}$, leading to a dangerous and synergistic accumulation of $\text{cGMP}$ that causes severe hypotension. \\
    \textbf{Reference knowledge:} Drug molecular structure (Nitrate functional group in Erythrityl tetranitrate) \\ \hline
  \end{tabular}
  \label{tab:limit}
\end{table}

\subsection{The Classification of Knowledge Sources}
\label{app:source}

As summarized in Table~\ref{tab:knowledge_source}, the DDI prediction is enhanced by introducing several types of knowledge sources. 
For both DrugBank and TWOSIDES, this includes drug SMILES, drug-target interactions, and drug-side effect information. A critical component of the knowledge source is the pharmacological categorization of known DDIs. 
Specifically, DDIs in DrugBank are divided into six distinct classes based on their pharmacokinetic and pharmacodynamic mechanisms, while DDIs in TWOSIDES are categorized into nine medical system-based classes.

\begin{table*}[h]
  \centering
  \small
  \caption{Knowledge source categorization for different datasets.}
  \vspace{-5px}
  \begin{tabular}{lp{3.5cm}p{11.5cm}}
    \hline
    \textbf{Dataset} & \textbf{Knowledge sources} & \textbf{Explanation}\\
    \hline
    \multirow{12}{*}{DrugBank} & Drug SMILES & String representations of drug chemical structures. \\
    & Drug target interactions& Binding relationships between drug molecules and their specific biological targets. \\
    & Drug side effects & Adverse or unintended physiological responses observed after the administration of drugs. \\
    & \multirow{2}{*}{Drug description} & Concise textual summary providing essential details about a medication's purpose, chemical composition, and safety profile. \\
    & DDI description & A brief introduction of drug-drug interaction types provided as choices in DDI prediction question. \\
    & Absorption DDIs& Interactions that alter the rate or extent of drug entry into the bloodstream. \\
    & Distribution DDIs& Interactions affecting the movement of drugs between the blood and various body tissues. \\
    & Metabolic DDIs& Interactions influencing the biochemical transformation of drugs. \\ 
    & Excretion DDIs & Interactions impacting the processes through which drugs and metabolites are removed from the body. \\
    & Synergistic DDIs& Interactions where the combined drug effect is greater than the sum of individual effects. \\
    & Antagonistic DDIs& Interactions where one drug reduces or counteracts the pharmacological effect of another. \\\hline
    \multirow{22}{*}{TWOSIDES} & Drug SMILES& String representations of drug chemical structures. \\
    & \multirow{2}{*}{Drug description} & Concise textual summary providing essential details about a medication's purpose, chemical composition, and safety profile. \\
    & DDI description & A brief introduction of drug-drug interaction types provided as choices in DDI prediction question. \\
    & Drug target interactions& Binding relationships between drug molecules and their specific biological targets.\\
    & Drug side effects& Adverse or unintended physiological responses observed after the administration of drugs. \\
    & Nervous System and Psychiatric Disorders& \multirow{2}{*}{Side effects affecting brain function, neurological processes, or mental health.} \\
    & Cardiovascular and Circulatory Disorders& \multirow{2}{*}{Adverse interaction effects impacting the heart, blood vessels, and systemic circulation.}\\
    & Respiratory Disorders& Side effects affecting the lungs and breathing processes caused by interacting drug pairs. \\
    & Gastrointestinal and Hepatic Disorders& \multirow{2}{*}{Drug interaction effects related to the digestive system and liver function.} \\
    & Renal and Metabolic Disorders& Adverse effects involving kidney function or systemic biochemical processes and regulation. \\
    & Immune, Hematologic, and Endocrine Disorders& \multirow{2}{*}{Interactions affecting the immune system, blood components, or hormone-producing glands.} \\
    & Musculoskeletal and Connective Tissue Disorders& \multirow{2}{*}{Side effects involving muscles, bones, joints, and associated supportive tissues.} \\
    & Dermatologic and Sensory Disorders& \multirow{2}{*}{Adverse reactions impacting the skin, hair, nails, or sensory organs.} \\
    & Neoplastic and Systemic Conditions& \multirow{2}{*}{Interactions related to tumor growth or generalized, whole-body physiological complications.} \\\hline
  \end{tabular}
  \label{tab:knowledge_source}
\end{table*}

\subsection{Implementation Details of Knowledge Retrieval by Expert Agents}
\label{app:retrieve}
Before expert agent analysis, the knowledge context is obtained through a retrieval process as follows:
\begin{itemize}[leftmargin=*]
\item A textual encoder transforms the DDI question and all knowledge entries within the assigned sources into embeddings. 
\item Relevance scores are determined via cosine similarity between the query embedding and each entry embedding.
\item We perform a top-k retrieval operation independently for each type of knowledge source assigned to the agent. This prevents a single dominant source from overshadowing other relevant but less frequent data types. The final context is the union of these retrieved subsets. 
\end{itemize}

\subsection{Algorithm of DDIAgents}
The algorithm of DDIAgents is shown in Algorithm~\ref{alg:ddiagents}, which includes the iteration of three critical stages: Expert agent instantiation, Dynamic context flow, Analysis and Decision making. 

\begin{algorithm}[ht]
\caption{DDIAgents for DDI Prediction through Multi-Agent System}
\label{alg:ddiagents}
\begin{algorithmic}[1] 
\Require DDI Question, Knowledge sources and Knowledge textual encoder

\State $t \leftarrow 1$
\While{True}
    \State // Stage (a): Expert agent instantiation
    \If{$t = 1$}
        \State Instantiate expert agents as predefined domain experts (Pharmacist, Pharmacokineticist and Pharmacologist)
    \Else
        \State Expert agent instantiation based on feedback of conclusion agent in $t-1$ iteration
    \EndIf

    \State // Stage (b): Dynamic context flow
    \For{each expert agent}
        \State Evaluate knowledge source assigned for the expert agent
    \EndFor

    \State // Stage (c): Analysis and decision making
    \State Identify allocated data source types from dynamic context flow
    \State For each expert, retrieve top-$k$ knowledge entries based on semantic similarity through knowledge textual encoder
    \For{each expert agent}
        \State Generate specific expert analysis
    \EndFor
    \State Synthesize findings from all agents

    \If{$\mathcal{O}_C^{t}$ requires refinement}
        \State Generate iteration guidance
        \State $t \leftarrow t + 1$ 
    \Else
        \State \textbf{break} 
    \EndIf
\EndWhile
\State \Return final answer and explanation

\end{algorithmic}
\end{algorithm}


\subsection{Statistics of Datasets}

We present the general statistics of datasets in experiments in Table~\ref{tab:statistics}. 
Here $\mathcal{D}$ represents the set of drugs, $\mathcal{R}$ denotes the set of interactions and $\mathcal{N}$ refers to the set of DDIs. 


\begin{table}[H]
  \centering
  \caption{Statistics of DDI prediction datasets in experiments.}
  \begin{tabular}{c|cc}
    \hline
    Dataset & DrugBank & TWOSIDES \\\hline
    $|\mathcal{D}_k|$ (Known drug set)  & 1020 & 292 \\
    $|\mathcal{D}_n|$ (New drug set) & 129 & 38 \\
    $|\mathcal{R}|$ (Relation types) & 86 & 209 \\
    $|\mathcal{N}_{\text{train}}|$ (Train DDIs) & 83272 & 11300 \\
    $|\mathcal{N}_{\text{test S1}}|$ (Test DDIs for S1 tasks) & 20031 & 3203 \\
    $|\mathcal{N}_{\text{test S2}}|$ (Test DDIs for S2 tasks) & 1298 & 194 \\ \hline
  \end{tabular}
  \label{tab:statistics}
\end{table}

\subsection{Evaluation Metrics}
\label{app:metric}
According to the evaluation metrics mentioned in Section~\ref{sec:exp_setup}, the evaluation metrics include F1-Score, accuracy for DrugBank: 
\begin{itemize}[leftmargin=*]
\item F1-Score~(Macro) $ = \frac{1}{||\mathcal{P}_D||}\sum_{p\in\mathcal{P}_D}\frac{2P_p \cdot R_p}{P_p + R_p}$, where $P_p$ and $R_p$ are the precision and recall for the interaction type $p$, respectively. 
\item Accuracy: the proportion of correctly predicted interaction types relative to the ground-truth interaction types.
\end{itemize} 

For TWOSIDES dataset, we use Hit@5 and NDCG@5 as evaluation metrics. Denote $\mathcal{I}_{1:5}$ as the recommended interaction for the given query drug pairs, $\mathcal{T}$ as the ground-truth interaction list and $\mathbb{I}$ as the indicator function. These metrics can be represented as follows:  
\begin{itemize}[leftmargin=*]
\item Hit@5 $=\mathbb{I}(|\mathcal{I}_{1:5} \cap \mathcal{T}|)$. 
\item NDCG@5 $=\frac{\sum_{i=1}^5 \mathbb{I}(\mathcal{I}_i \in T)/\log_2(i+1)}{\sum_{i=1}^{\min(|T|, 5)}1/\log_2(i+1)}$ .
\end{itemize} 

\subsection{Baseline Methods}

Here we present a more detailed introduction of baseline methods compared in experiments: 

\begin{itemize}[leftmargin=*]
\item MLP~\citep{rogers2010extended} is a classic supervised learning approach that utilizes a multi-layer perceptron to map drug chemical fingerprints to specific interaction types between drug pairs.
\item MSTE~\citep{yao2022effective} specially designs a knowledge graph embedding scoring function to model complex relations in DDI prediction problems.
\item Decagon~\citep{zitnik2018modeling} is a representative GNN-based approach that performs multi-relational link prediction by extracting and aggregating information from biomedical networks.
\item EmerGNN~\citep{zhang2023emerging} is a specialized GNN that utilizes a flow-based architecture to identify critical paths within biomedical networks, specifically optimized for predicting interactions involving emerging drugs.
\item Single LLM is a straightforward method that directly uses LLMs to answer textual DDI prediction questions without adding any supplementary information. In our experiments, we implement this baseline using Qwen2.5-7B as the backbone model.
\item TIGER~\citep{su2024dual} is a hybrid architecture featuring a dual-channel graph transformer, which simultaneously captures the internal structural information of drug molecular graphs and the global topological features of biomedical networks.
\item TextDDI~\citep{zhu2023learning} is the first framework to leverage the zero-shot capabilities of LLMs for DDI prediction, utilizing textual drug descriptions and interaction histories.
\item DDI-GPT~\citep{xu2024ddi} captures relevant information of query drugs from biomedical networks and uses biomedical large language model to enhance DDI prediction performance. 
\item CBR-DDI~\citep{liu2025case} adapts the "Case-Based Reasoning" paradigm to the LLM setting, enabling the LLM model to retrieve and learn from past DDI instances to solve new DDI queries.
\item K-path~\citep{abdullahi2025k} extracts critical paths from biomedical networks as important background to enhance the LLM prediction performance on DDI prediction problem. 
\item Reflexion~\citep{shinn2023reflexion} is a reinforcement learning-inspired architecture that enables single agent to improve its performance through self-reflection, where it critiques past mistakes and stores them in a verbal memory to inform more accurate future attempts.
\item Debate~\citep{du2023improving} is a multi-agent collaboration protocol where specialized agents iteratively critique and refine each other's responses until a consensus is reached.
\item AgentVerse~\citep{chen2023agentverse} is a dynamic framework that adjusts the composition of an agent group based on the specific progress and complexity of the problem-solving task.
\item MDAgents~\citep{kim2024mdagents} is a multi-agent system specifically tailored for medical decision-making, employing adaptive protocols that scale in complexity based on the perceived difficulty of the clinical case.
\end{itemize} 

\subsection{Prompt used in Experiments}
\label{app:prompt}

In this section, we provide the prompt templates of different agents in DDIAgents in Table~\ref{tab:pmpt1}-Table~\ref{tab:pmpt4}. 

\begin{table}[h]
  \centering
  \small
  \caption{System prompt for planner in expert agent instantiation.}
  \vspace{-5px}
  \begin{tabular}{p{8cm}}
  \hline
  \texttt{You are a medical expert who specializes in categorizing a specific drug-drug interaction prediction problem into specific areas of medicine. }\\
  \texttt{You need to complete the following steps:} \\
  \texttt{1. Carefully read the drug-drug interaction prediction problem with candidate choices and feedbacks from last iteration round.} \\
  \texttt{2. Based on the problem, provide three experts that are suitable to answer the question from different perspectives.} \\
  \texttt{3. Provide three expert names and use one sentence to describe their roles, respectively.} \\
  \texttt{You should output in exactly the same format as: (1) [expert1 name]: [one sentence describes the role of expert1]. (2) [expert2 name]: [one sentence describes the role of expert2]. (3) [expert3 name]: [one sentence describes the role of expert3]. }\\
  \texttt{Question: \{\{Question\}\}}\\
  \texttt{Candidate interaction types: \{\{Candidate DDI types\}\}}\\
  \texttt{Analysis in last round: \{\{Feedback from conclusion agent\}\}}\\
  \texttt{Please provide your split of three experts: }\\
  \hline
  \end{tabular}
  \label{tab:pmpt1}
\end{table}

\begin{table}[h]
  \centering
  \small
  \caption{System prompt for planner in dynamic context flow~(for DrugBank dataset).}
  \vspace{-5px}
  \begin{tabular}{p{8cm}}
  \hline
  \texttt{You are a medical librarian who specializes in managing medical knowledge sources for medical experts. }\\
  \texttt{Given the medical expert name, his expertise, drug-drug interaction prediction problem and knowledge sources, please select the relevant information needed. } \\
  \texttt{Expert name: \{\{Expert name\}\}. Expertise: \{\{Expertise\}\}. } \\
  \texttt{Question: \{\{Question\}\}}\\
  \texttt{Candidate interaction types: \{\{Candidate DDI types\}\}}\\
  \texttt{Knowledge sources: } \\
  \texttt{A. Absorption DDIs: an absorption DDI occurs when one drug alters the rate or extent of absorption and bioavailability of a second drug, leading to a decrease in its concentration and potential loss of efficacy or an increase in its concentration and potential worsening of adverse effects.} \\
  \texttt{B. Distribution DDIs: a distribution DDI occurs when one drug alters the distribution of a second drug throughout the body, typically by affecting plasma protein binding or transport, leading to changes in the free serum concentration of Drug2, which may result in altered efficacy or toxicity.} \\
  \texttt{C. Metabolic DDIs: a Metabolic DDI occurs when one drug alters the rate of metabolism of a second drug, typically via enzyme inhibition or induction, leading to either a change in Drug2's serum concentration or an alteration in the concentration of its active or inactive metabolites.} \\
  \texttt{D. Excretion DDIs: An Excretion DDI occurs when one drug alters the rate of renal or biliary excretion of a second drug, leading to either an increase in the elimination rate and potential loss of Drug2 efficacy or a decrease in the elimination rate and potential Drug2 toxicity due to elevated serum levels.} \\
  \texttt{E. Synergistic DDIs: a synergistic DDI is an interaction where two drugs combined produce an effect (therapeutic or adverse) that is greater than the sum of the effects of each drug given alone, typically resulting in an increase in the intensity of specific pharmacodynamic activities, such as toxicity or efficacy. } \\
  \texttt{F. Antagonistic DDIs: an antagonistic DDI is an interaction where one drug opposes the effect of a second drug by competing for the same receptor or having counteracting physiological effects, resulting in a decrease in the therapeutic efficacy or an attenuation of the adverse effects of Drug2.} \\
  \texttt{G. SMILES of two drugs. } \\
  \texttt{H. Drug targets: the targets two drugs both bind.} \\
  \texttt{I. Drug side effects: the side effects two drugs both cause.} \\
  \texttt{Please ONLY output your choices with letter sequences (from A-I) and separate them with spaces. } \\
  \hline
  \end{tabular}
  \label{tab:pmpt2}
\end{table}

\begin{table}[h]
  \centering
  \small
  \caption{System prompt for expert agent analysis.}
  \vspace{-5px}
  \begin{tabular}{p{8cm}}
  \hline
  \texttt{You are a \{\{Expert name\}\} who specializes in \{\{Agent expertise\}\}. }\\
  \texttt{You are provided with: } \\
  \texttt{- Two drugs and their descriptions.} \\
  \texttt{- Candidate interaction types predicted by a deep learning model, each with a short description.} \\
  \texttt{- Relevant knowledge sources as a reference.} \\
  \texttt{- Analysis summary of different agents in the last round.} \\
  \texttt{Note: }\\
  \texttt{- You MUST give your answer in English, MUST not contain characters in other language. }\\
  \texttt{- The interaction could be from Drug 1 to Drug 2, or from Drug 2 to Drug 1.}\\
  \texttt{Please choose the most plausible interaction type based on the provided information and your medical knowledge. }\\
  \texttt{Input format: }
  \texttt{Question: What is the drug-drug interaction type between [Drug 1] and [Drug 2]? }\\
  \texttt{[Drug 1]: [Description of Drug 1]; [Drug 2]: [Description of Drug 2].}\\
  \texttt{Candidate interaction types:}\\
  \texttt{- A. [Interaction Type A]: [Description of Interaction Type A]} \\
  \texttt{- B. [Interaction Type B]: [Description of Interaction Type B]} \\
  \texttt{...} \\
  \texttt{Reference knowledge sources: ...}\\
  \texttt{The answer of other agents in the last round: ...}\\
  \texttt{Output Format: }\\
  \texttt{Interaction type A/B/C/D/E. }\\
  \texttt{[Your analysis to explain your choice. ]}\\
  \texttt{Now answer this question: ...} \\
  \hline
  \end{tabular}
  \label{tab:pmpt3}
\end{table}

\begin{table}[h]
  \centering
  \small
  \caption{System prompt for decision making.}
  \vspace{-5px}
  \begin{tabular}{p{8cm}}
  \hline
  \texttt{You are a medical expert specializing in pharmacology and drug safety. }\\
  \texttt{Your task is to predict the most likely drug-drug interaction (DDI) between a given pair of drugs. } \\
  \texttt{You are provided with the prediction and information from multiple experts. } \\
  \texttt{Please take their answer into comprehensive consideration and judge whether the evidence provided is enough to obtain a conclusion. } \\
  \texttt{You are provided with: } \\
  \texttt{- Two drugs and their descriptions. } \\
  \texttt{- Candidate interaction types predicted by a deep learning model, each with a short description.}\\
  \texttt{- The answer of other agents for reference. }\\
  \texttt{Note: }\\
  \texttt{- You MUST give your answer in English, MUST not contain characters in other language. }\\
  \texttt{- If there are contradictions among the answer of different agents, it is encouraged to require a new round of analysis for different experts.  }\\
  \texttt{Input format: }
  \texttt{Question: What is the drug-drug interaction type between [Drug 1] and [Drug 2]? }\\
  \texttt{[Drug 1]: [Description of Drug 1]; [Drug 2]: [Description of Drug 2].}\\
  \texttt{Candidate interaction types:}\\
  \texttt{- A. [Interaction Type A]: [Description of Interaction Type A]} \\
  \texttt{- B. [Interaction Type B]: [Description of Interaction Type B]} \\
  \texttt{...} \\
  \texttt{The answer of other agents: ...}\\
  \texttt{Output must strictly follow the format below.}\\
  \texttt{If you think the evidence provided is enough, the output format MUST be: }\\
  \texttt{Yes. Interaction type A/B/C/D/E. }\\
  \texttt{[Your analysis to explain your choice. ]}\\
  \texttt{If you think the evidence provided is not enough, the output format MUST be:}\\
  \texttt{No. [One sentence to summarize the answer of each expert agent, respectively.]}\\
  \texttt{[Suggestion for further analysis on the question. ] }\\
  \texttt{Now answer this question: ...} \\
  \hline
  \end{tabular}
  \label{tab:pmpt4}
\end{table}




\clearpage

\section{More Experimental Results}
\label{sec:more_exp}

\subsection{Experimental Results on S0 Setting}
\label{sec:s0}

Table~\ref{tab:s0_performance} details the performance of various methods within the S0 tasks, which focuses on predicting interactions between known drugs. 
While the proposed DDIAgents maintain strong performance, the performance gap between models is less pronounced in S0 than in S1 and S2 tasks. 
This suggests that traditional methods can effectively leverage the rich interaction history of known drugs, making the prediction task less challenging than in zero-shot or cold-start scenarios.

\subsection{Experimental Results on TWOSIDES Dataset with Traditional Evaluation Setting}
\label{app:two}

To provide a comprehensive comparison with existing literature, we report the performance of DDIAgents on the TWOSIDES dataset using the traditional binary classification setting. 
While we maintain that the ranking-based metrics (Hit@5 and NDCG@5) introduced in Section~\ref{sec:exp_setup} better reflect clinical realities by addressing the multi-label nature of drug-drug interactions, these supplementary results in Table~\ref{tab:two_old} serve as a benchmark against standard methodologies. 
Here we use the following two evaluation metrics:
\begin{itemize}[leftmargin=*]
\item ROC-AUC $ = \sum_{k = 1}^n TP_k \Delta FP_k$ measures the area curve of receiver operating characteristics. 
\item Accuracy: the proportion of correctly predicted DDIs for each DDI type. 
\end{itemize}
As shown, DDIAgents consistently outperforms all baseline methods across both S1 and S2 tasks, demonstrating its robustness when evaluated under traditional binary classification setting.

\begin{table}[h]
  \centering
  \setlength{\tabcolsep}{3pt}
  \caption{Performance of representative methods on TWOSIDES with traditional setting. The best and second-best results are highlighted in bold and underline, respectively.}
  \begin{tabular}{l|cccc}
    \hline
    &\multicolumn{4}{c}{\textbf{TWOSIDES}} \\
    Method & \multicolumn{2}{c}{S1}  & \multicolumn{2}{c}{S2}   \\
    &ROC-AUC&Acc&ROC-AUC&Acc \\\hline
    MLP &74.72$\pm$1.39&66.54$\pm$0.93&54.82$\pm$1.72&51.64$\pm$2.29\\
    TIGER&81.59$\pm$0.82&72.48$\pm$1.36&61.48$\pm$0.91&52.77$\pm$0.55\\ 
    DDI-GPT&\underline{86.50$\pm$0.89}&78.25$\pm$0.40&\underline{64.57$\pm$1.53}&\underline{61.26$\pm$1.27}\\ 
    Reflexion &83.07$\pm$0.61&75.41$\pm$0.49&60.71$\pm$0.88&56.44$\pm$1.21 \\
    MDAgents&85.16$\pm$0.69&\underline{78.39$\pm$0.71}& 63.08$\pm$1.98 & 60.74$\pm$1.49 \\ 
    DDIAgents&\textbf{87.36$\pm$0.41}&\textbf{80.09$\pm$0.59}& \textbf{66.41$\pm$1.02} & \textbf{62.77$\pm$0.78} \\ \hline
  \end{tabular}
  \label{tab:two_old}
\end{table}

\subsection{Candidate Retrieval Performance of the Traditional Model}
\label{app:trad}

To enable efficient agent-based DDI prediction, we first train a traditional model as a \emph{candidate retriever} that filters a small set of highly probable interaction types for each drug pair. 
Specifically, we keep the top-5 candidates for DrugBank and the top-20 candidates for TWOSIDES, aiming to maintain a high upper bound on recall; otherwise, the downstream agents would be fundamentally limited by missed candidates.

Table~\ref{tab:old4agent} reports the retrieval statistics of this traditional model. 
Overall, the model can successfully narrow the DDI candidate choice while retaining most true interaction types among the retrieved candidates. 
However, its prediction quality remains limited, as reflected by substantially lower accuracy/Hit@5 rates.
For instance, on DrugBank, the retriever achieves high Recall@5 (93.92\% in S1 and 89.46\% in S2), yet its accuracy is only 50.68\% (S1) and 33.47\% (S2), indicating that high-recall filtering alone is insufficient for accurate DDI prediction.
These results support our design choice: the traditional model serves to maximize recall and reduce the candidate space, while the performance gains of multi-agent systems mainly come from the subsequent agent-based analysis that improves precision through mechanism-aware reasoning.


\begin{table}[h]
  \centering
  \setlength{\tabcolsep}{3pt}
  \caption{Candidate retrieval performance of the traditional model used for agent-based DDI prediction. Here ``(agt)'' refers to the result of DDIAgents. }
  \begin{tabular}{cccccc}
    \hline
    \multicolumn{6}{c}{\textbf{DrugBank}} \\
    \multicolumn{3}{c}{S1}  & \multicolumn{3}{c}{S2}   \\
    Acc&Acc(agt)&Recall@5&Acc&Acc(agt)&Recall@5 \\\hline
    50.68&55.75&93.92&33.47&38.21&89.46\\
    \hline \hline
    \multicolumn{6}{c}{\textbf{TWOSIDES}} \\
    \multicolumn{3}{c}{S1}  & \multicolumn{3}{c}{S2} \\
    Hit@5&Hit@5(agt)&Hit@20&Hit@5&Hit@5(agt)&Hit@20 \\\hline
    43.90&53.94&78.23&33.51&42.26&68.56\\ \hline
  \end{tabular}
  \label{tab:old4agent}
\end{table}

\subsection{Ablation Study on Iteration Round of DDIAgents}
\begin{figure}[h]
	\centering
	\begin{subfigure}{0.23\textwidth}
		\centering
		\includegraphics[width=\linewidth]{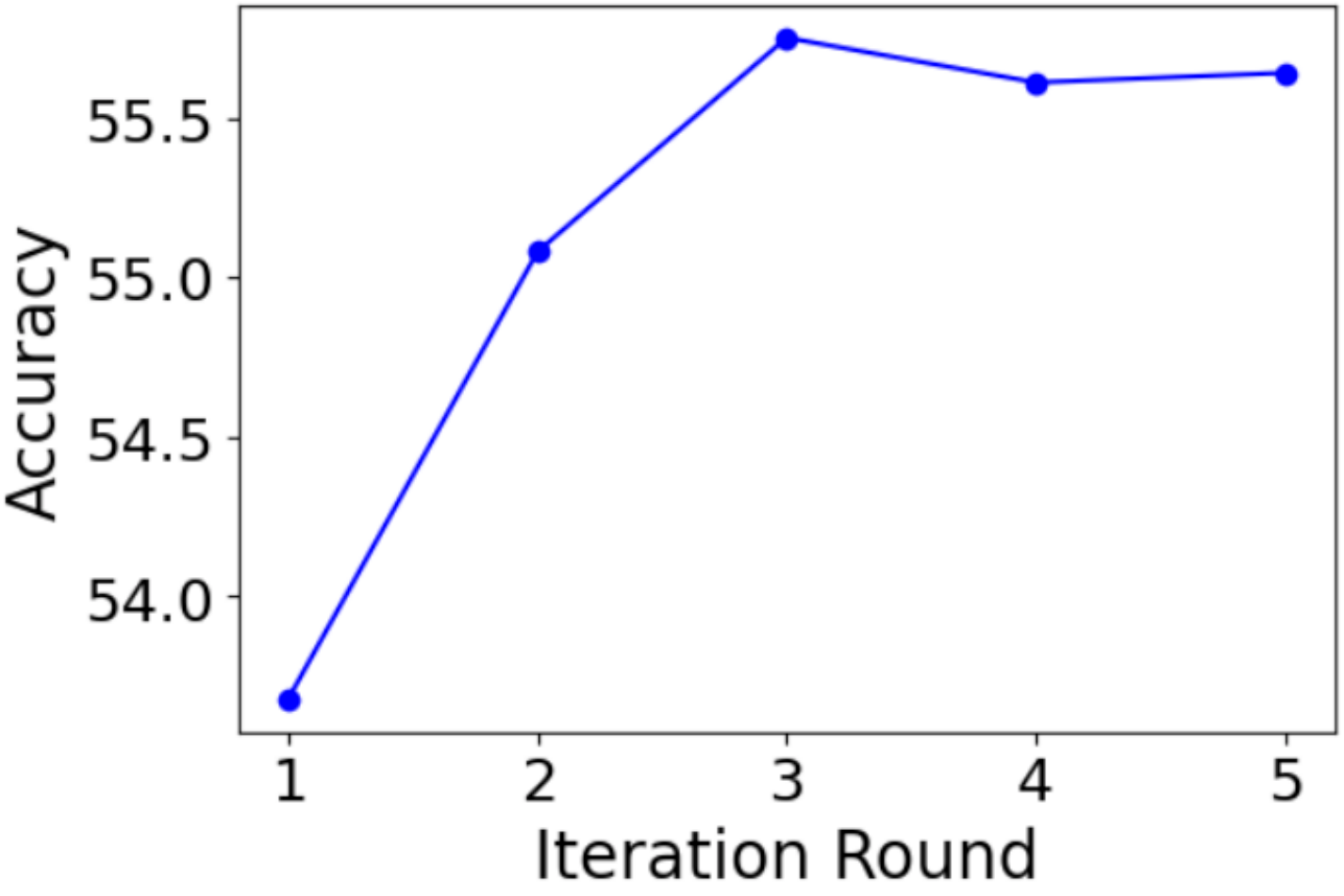}
		\caption{DrugBank, S1 tasks}
	\end{subfigure}
	\hfill 
	\begin{subfigure}{0.23\textwidth}
		\centering
		\includegraphics[width=\linewidth]{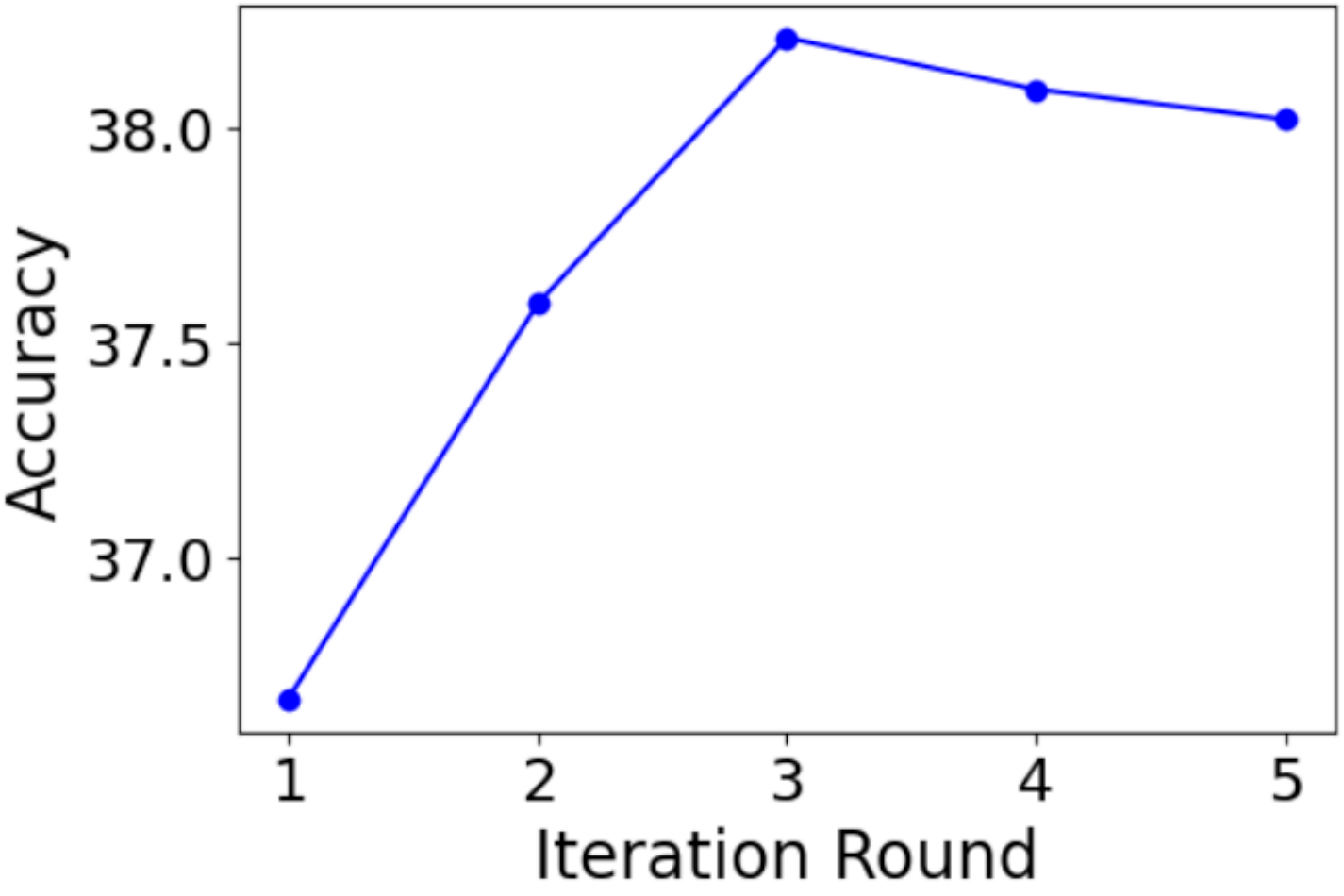}
		\caption{DrugBank, S2 tasks}
	\end{subfigure}
	\hfill 
	\begin{subfigure}{0.227\textwidth}
		\centering
		\includegraphics[width=\linewidth]{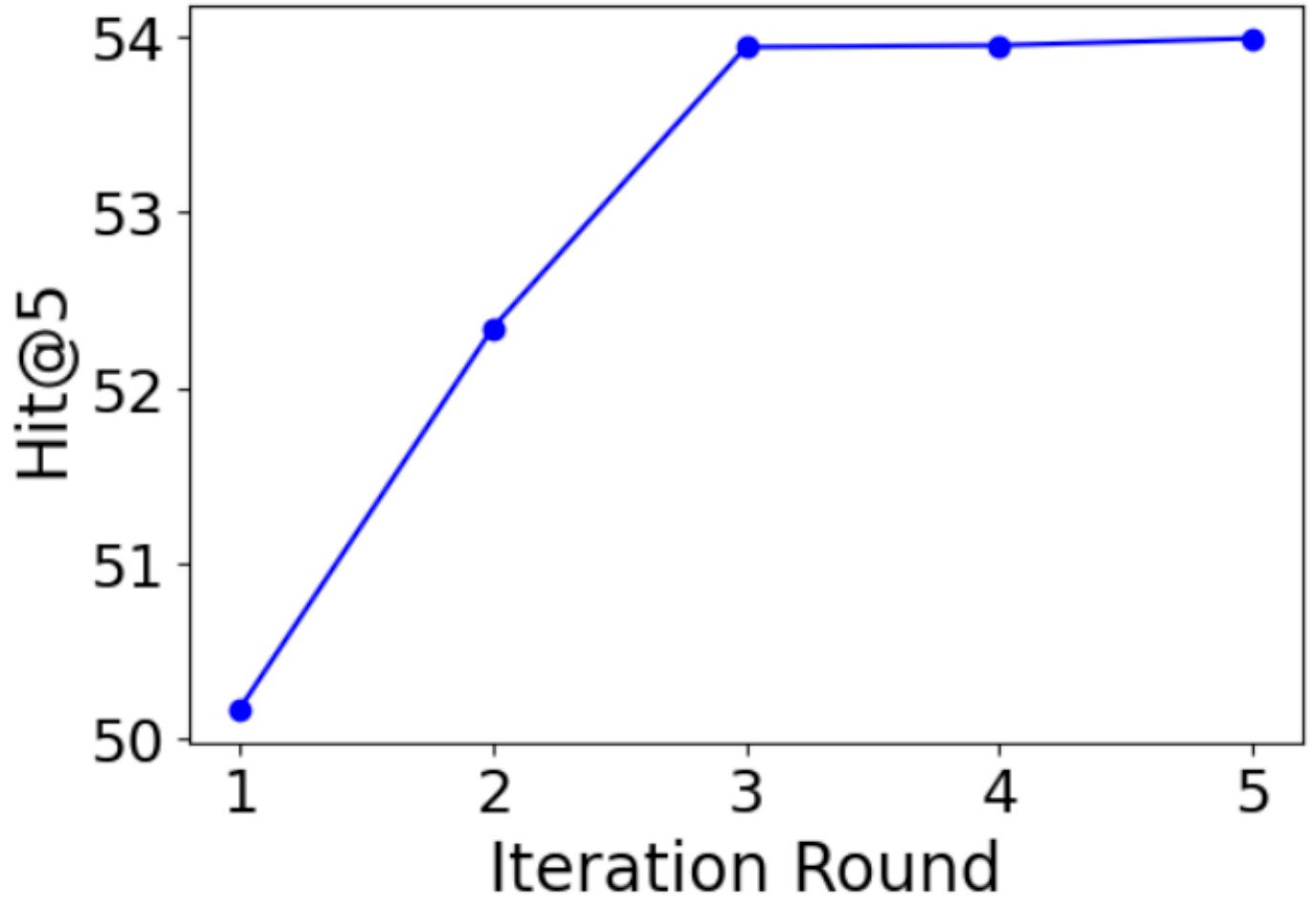}
		\caption{TWOSIDES, S1 tasks}
	\end{subfigure}
	\hfill 
	\begin{subfigure}{0.227\textwidth}
		\centering
		\includegraphics[width=\linewidth]{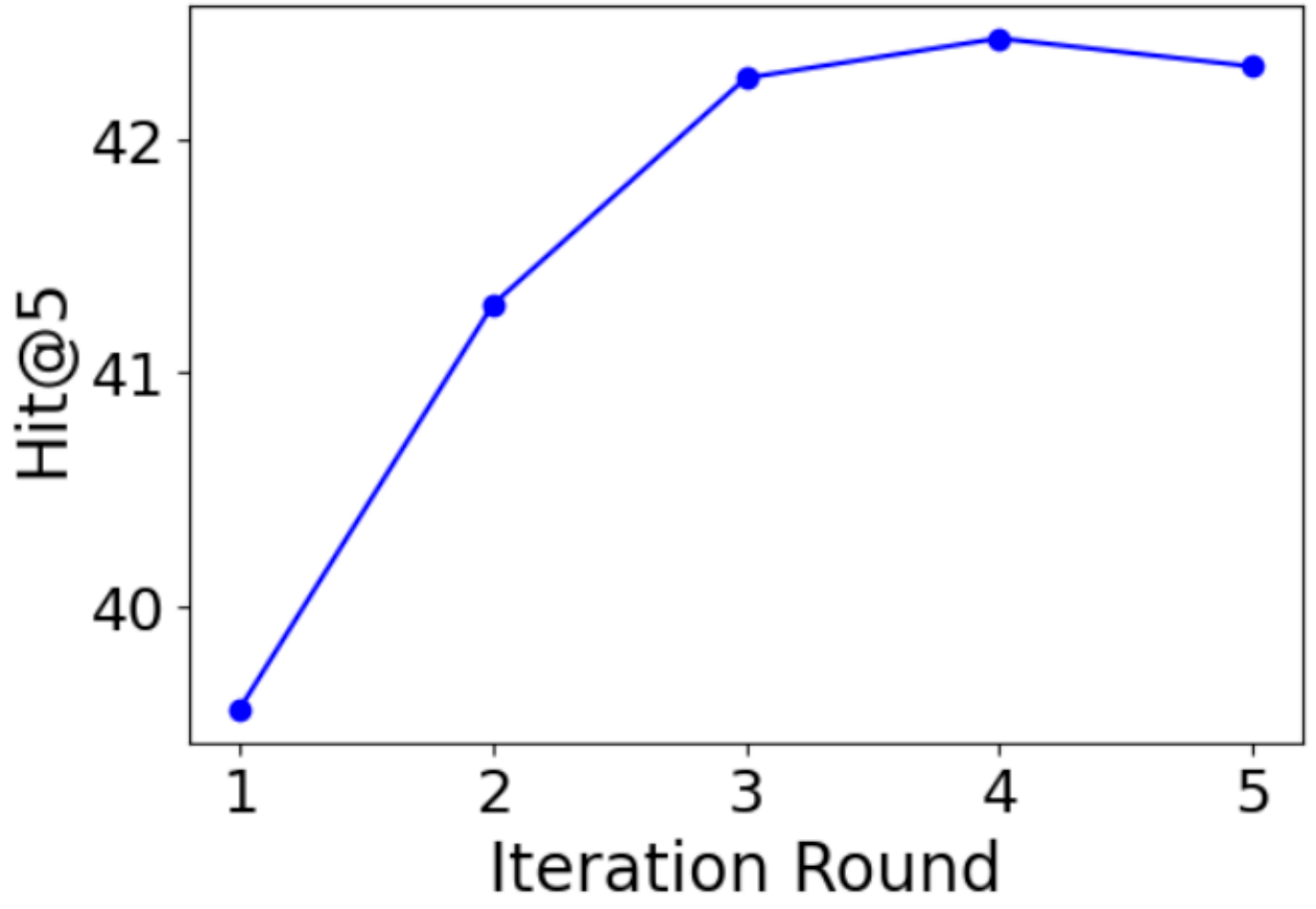}
		\caption{TWOSIDES, S2 tasks}
	\end{subfigure}
	\vspace{-7px}
	\caption{Varying the iteration round of DDIAgents. }
	\label{fig:iteration}
\end{figure}

Here we present the results varying the iterative analysis round $t$ in Figure~\ref{fig:iteration} to assess its impact on model efficacy. 
For initial analysis rounds ($t \leq 3$), the performance of DDIAgents consistently improves as $t$ increases, a gradual enhancement that strongly underscores the necessity of the iterative analysis process for effective DDI prediction. 
For analysis rounds extending beyond $t=3$, performance across both datasets stabilizes, indicating that the model achieves peak efficacy within a few iterations.

\subsection{Prediction Performance Using Large-scale Frontier Model}
\label{app:commercial}

Table~\ref{tab:gpt_performance} reports the S2 results of DDIAgents when replacing the backbone LLM with a large-scale frontier model~(GPT-4o). 
On DrugBank dataset, GPT-4o yields a clear gain in F1 score~(18.17$\rightarrow$20.75), suggesting that a stronger model particularly improves performance on the harder and less frequent DDI types captured by F1 score. 
On TWOSIDES, GPT-4o brings consistent improvements on both ranking metrics, which further highlights the prediction potential of DDIAgents under S2 tasks.


\subsection{More DDI prediction cases}
\label{app:case}

In this section, we provide additional DDI prediction case studies for DDIAgents using Qwen 2.5 7B as the backbone LLM. 
Table~\ref{app:example1} reports the same drug pair as the case illustrated in Figure~\ref{fig:case} (Section~\ref{exp:case}). 
While DDIAgents with GPT-4o correctly predicts the ground-truth interaction in Section~\ref{exp:case}, the Qwen 2.5 7B backbone produces an incorrect prediction after a single iteration. 
This comparison suggests that stronger backbone reasoning can translate into improved overall performance for DDIAgents. 
In addition, Table~\ref{app:example2} presents a case where DDIAgents with Qwen 2.5 7B arrives at the correct prediction after two iterations, highlighting the benefit of iterative refinement even with a smaller backbone model.

\subsection{The Statistics of Newly Instantiated Agents}

To characterize the dynamics of expert agent instantiation, we summarize the most frequently created new expert agents and their predictive utility on DrugBank S2 tasks in Table~\ref{tab:newagent}. 
The total number reflects how often DDIAgents identifies a mechanism-specific reasoning gap that is not well covered by the default experts and therefore requests a specialized perspective.

Overall, ``Cardiologist'', ``Neurologist'', and ``Toxicologist'' are instantiated most often, suggesting that a substantial portion of hard cases in DrugBank S2 require specialized clinical safety assessment and adverse-event reasoning beyond general pharmacology. 
Notably, instantiation frequency does not always align with per-agent accuracy: ``Endocrinologist'' and ``Gastroenterologist'' achieve higher accuracy despite being created less often. 
This pattern indicates that certain expert perspectives are high-yield when triggered (i.e., they provide decisive evidence for a narrower subset of queries), whereas other experts are broad-coverage but face more ambiguous cases.

These statistics provide empirical evidence that DDIAgents benefits from dynamic expert instantiation: rather than relying on a fixed agent pool, it adaptively composes the experts needed to match the interaction-dependent evidence required by each query.


\begin{table}[h]
  \centering
  \setlength{\tabcolsep}{3pt}
  \caption{The statistics of newly instantiated agents. }
  \begin{tabular}{ccc}
    \hline
    Expert name & Total number & Accuracy \\
    Cardiologist & 142 & 40.14 \\
    Neurologist & 71 & 36.62 \\
    Toxicologist & 55 & 41.82 \\
    Pulmonologist & 53 & 33.96 \\
    Endocrinologist & 52 & 44.23 \\
    Hepatologist & 44 & 38.64 \\
    Oncologist & 41 & 39.02 \\
    Gastroenterologist & 39 & 43.59 \\
    Nephrologist & 36 & 36.11 \\
    Pharmacogeneticist & 31 & 32.26 \\
    \hline
  \end{tabular}
  \label{tab:newagent}
\end{table}

\subsection{Case Study: the Correlation between Expert and Knowledge Source in Dynamic Context Flow}

\begin{figure}[h]
	\centering
	\includegraphics[width=0.47\textwidth]{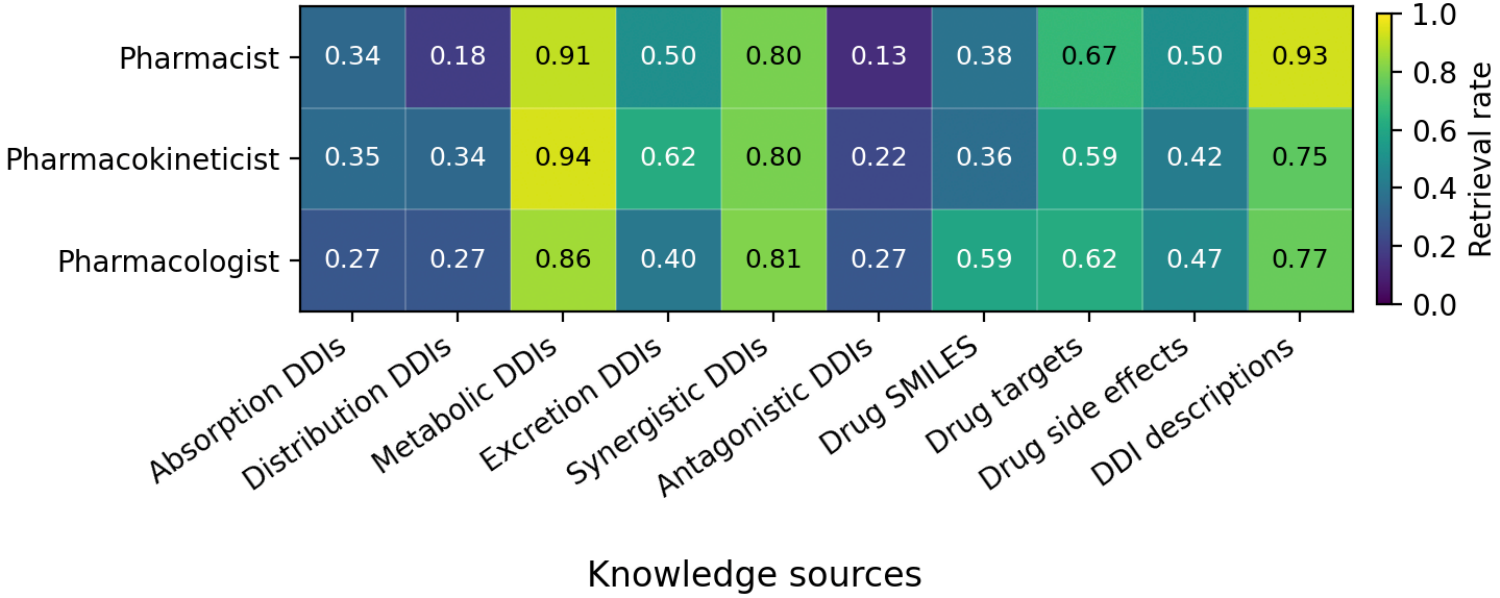}
  \vspace{-5px}
	\caption{Knowledge retrieval rates across expert agents and knowledge sources on DrugBank S2 tasks. }
	\label{fig:case_flow}
\end{figure}

We conduct a case study to analyze the context flow of DDIAgents on the DrugBank dataset under S2 tasks (as shown in Figure~\ref{fig:case_flow}). 
For clarity, we illustrate three predefined expert agents, noting that the instantiated expert profiles can vary across iterations.
The retrieval rates of each expert over different knowledge sources are illustrated.

Overall, Metabolic DDIs, Synergistic DDIs, and DDI descriptions are retrieved most frequently across experts, suggesting that these sources provide particularly informative evidence for accurate DDI prediction.
Moreover, the experts exhibit distinct retrieval preferences consistent with their functional roles. 
The Pharmacist emphasizes clinically oriented evidence (e.g., drug side effects, drug targets, and textual DDI descriptions), 
the Pharmacokineticist focuses on ADME-related sources (Absorption, Distribution, Metabolic and Excretion DDIs), 
and the Pharmacologist places relatively more weight on molecular structure signals (e.g., Drug SMILES) for chemical mechanism analysis.


\begin{table*}[h]
  \centering
  \setlength{\tabcolsep}{7pt}
  \caption{Performance comparison between different DDI prediction methods on S0 tasks. The best and second-best results are highlighted in bold and underline, respectively.}
  \begin{tabular}{c|l|cc|cc}
    \hline
    &&\multicolumn{2}{c|}{\textbf{DrugBank}}&\multicolumn{2}{c}{\textbf{TWOSIDES}} \\
    Type &Method & \multicolumn{2}{c|}{S0}  & \multicolumn{2}{c}{S0}   \\
    &&Acc&F1&Hit@5&NDCG@5 \\\hline
    Feature-based & MLP &84.30$\pm$1.13&19.10$\pm$2.65&66.54$\pm$1.39&30.75$\pm$1.21\\\hline
    \multirow{4}{*}{Graph-based}&Decagon&90.61$\pm$0.71&85.65$\pm$1.10&54.07$\pm$0.47&21.10$\pm$1.06\\
    &EmerGNN&96.71$\pm$0.11&94.12$\pm$0.39&68.71$\pm$0.29&33.09$\pm$0.31\\
    &MSTE&92.17$\pm$0.27&71.30$\pm$2.74&30.91$\pm$0.07&12.91$\pm$0.04\\
    &TIGER&95.85$\pm$0.83&92.31$\pm$0.31&62.84$\pm$0.77&29.51$\pm$0.41\\ \hline
    \multirow{5}{*}{LLM-based}& Single LLM &9.17$\pm$0.26&4.79$\pm$0.19& 20.44$\pm$0.19 & 9.82$\pm$0.12 \\
    &K-Path&92.49$\pm$2.27&89.86$\pm$2.71&55.39$\pm$1.81&24.14$\pm$1.36\\ 
    &CBR-DDI&\underline{96.84$\pm$0.62}&\underline{94.92$\pm$0.76}&69.17$\pm$0.52&\underline{35.40$\pm$0.16}\\ 
    &TextDDI&96.19$\pm$0.48&91.54$\pm$0.33&57.09$\pm$0.77&28.51$\pm$0.39\\ 
    &DDI-GPT&96.63$\pm$1.31&93.32$\pm$0.88&68.16$\pm$1.48&34.51$\pm$0.92\\ \hline
    \multirow{5}{*}{Agent-based}& Reflexion &92.04$\pm$1.82&88.39$\pm$2.04&58.45$\pm$1.29&26.86$\pm$0.77 \\
    &Debate&94.41$\pm$0.92&93.12$\pm$1.36&63.11$\pm$0.68&32.10$\pm$0.99\\ 
    &AgentVerse&95.77$\pm$1.41&94.04$\pm$1.27&67.08$\pm$0.82&33.09$\pm$0.55\\ 
    &MDAgents&96.33$\pm$0.97&94.19$\pm$0.72& \underline{69.63$\pm$1.22} & 33.67$\pm$1.01 \\ \cline{2-5}
    &DDIAgents&\textbf{96.95$\pm$0.84}&\textbf{95.02$\pm$1.09}& \textbf{71.57$\pm$0.72} & \textbf{35.82$\pm$0.44} \\ \hline
  \end{tabular}
  \label{tab:s0_performance}
\end{table*}

\begin{table*}[h]
  \centering
  \caption{Performance of DDIAgents using GPT-4o as the backbone LLM.}
  \vspace{-5px}
  \begin{tabular}{l|cc|cc}
    \hline
    &\multicolumn{2}{c|}{\textbf{DrugBank}}&\multicolumn{2}{c}{\textbf{TWOSIDES}} \\
    Method & \multicolumn{2}{c|}{S2}  & \multicolumn{2}{c}{S2}   \\
    &Acc&F1&Hit@5&NDCG@5 \\\hline
    DDIAgents~(Qwen 2.5 7B) &38.21$\pm$1.03 & 18.17$\pm$0.51 & 42.26$\pm$0.92 & 14.08$\pm$0.15\\
    DDIAgents~(GPT-4o) &38.34$\pm$0.62&20.75$\pm$0.69&44.08$\pm$0.79&14.90$\pm$0.24\\ \hline
  \end{tabular}
  \label{tab:gpt_performance}
\end{table*}

\begin{table*}[h]
  \centering
  \caption{The DDI prediction case by DDIAgents using Qwen 2.5 7B as the backbone LLM~(The same DDI prediction question as illustrated in Section~\ref{exp:case}). }
  \vspace{-5px}
  \begin{tabular}{p{2.5cm}|p{14.5cm}}
  \hline
  DDI prediction question& What is the drug-drug interaction type between Pirfenidone and Teriflunomide? \newline A. Decreased metabolism B. Increased risk or severity of adverse effects C. Increased serum concentration \newline D. Decreased serum concentration E. Decreased therapeutic efficacy  \\\hline
  Pharmacist \newline (Iteration 1)& \textbf{Assigned knowledge sources:} Metabolic DDIs, Excretion DDIs, Synergistic DDIs and Drug side effects. \newline  \textbf{Answer: Interaction type A.} Pirfenidone is a small molecule drug with multiple anti-fibrotic properties, while Teriflunomide is a disease-modifying treatment for multiple sclerosis. The metabolism of Teriflunomide can be decreased when combined with Pirfenidone, this suggests that Pirfenidone may inhibit enzymes that break down Teriflunomide, leading to elevated plasma levels and a higher risk of toxicity. This is a common scenario in drug interactions, especially in patients with multiple chronic conditions where polypharmacy is prevalent. Monitoring for adverse effects and potential dose adjustments would be necessary to manage this interaction effectively. 
  \\\hline
  Pharmacokineticist \newline (Iteration 1)& \textbf{Assigned knowledge sources:} Metabolic DDIs, Excretion DDIs, Synergistic DDIs and Antagonistic DDIs.  \newline \textbf{Answer: Interaction type A.} Pirfenidone is a small molecule with multiple anti-fibrotic properties, and Teriflunomide is a novel disease-modifying treatment for multiple sclerosis. Based on the candidate interaction types, the most plausible interaction is that the metabolism of Teriflunomide can be decreased when combined with Pirfenidone. This interaction indicates that Pirfenidone inhibits enzymes that break down Teriflunomide, leading to elevated plasma levels and a higher risk of toxicity. This aligns with the deep learning model's prediction, which has a high probability score, suggesting a strong likelihood of this interaction occurring. Therefore, monitoring for increased toxicity and potential dose adjustments would be necessary if these drugs are co-administered.
  \\\hline
  Pharmacologist \newline (Iteration 1)& \textbf{Assigned knowledge sources:} Metabolic DDIs, Excretion DDIs, Synergistic DDIs, Antagonistic DDIs, Drug targets and Drug side effects.  \newline \textbf{Answer: Interaction type C.} Pirfenidone is a drug with multiple anti-fibrotic properties, while Teriflunomide is a novel disease-modifying treatment for multiple sclerosis (MS) that is in phase III development. Pirfenidone may inhibit enzymes that break down Teriflunomide, leading to elevated plasma levels and a higher risk of toxicity. This is a plausible interaction given that both drugs are in different therapeutic areas and their mechanisms of action do not directly overlap in terms of receptors or pathways. Therefore, the most plausible interaction type is A, where the metabolism of Teriflunomide is decreased when combined with Pirfenidone.
  \\\hline
  Conclusion agent \newline (Iteration 1)&  \textbf{Answer: Yes. Interaction type C.} The agents all agreed that the metabolism of Teriflunomide could be inhibited by Pirfenidone, leading to increased plasma levels and a higher risk of toxicity. Therefore, this interaction type is the most likely. 
  \\\hline
  \end{tabular}
  \label{app:example1}
\end{table*}

\begin{table*}[h]
  \centering
  \caption{A DDI prediction case by DDIAgents that different agents achieve consensus in two iterations.}
  \vspace{-5px}
  \begin{tabular}{p{2.5cm}|p{14.5cm}}
  \hline
  DDI prediction question& What is the drug-drug interaction type between Riociguat and Macitentan? \newline A. Decreased serum concentration B. Increased anticoagulant activities \newline C. Increased risk or severity of adverse effects  D. Increased serum concentration E. Increased hypotensive activities \\\hline
  Pharmacist \newline (Iteration 1)& \textbf{Assigned knowledge sources:} Absorption DDIs, Metabolic DDIs and SMILES of two drugs.  \newline  \textbf{Answer: Interaction type E.} Riociguat, a PDE5 inhibitor that increases cGMP levels, and Macitentan, a VEGFR-1 and VEGFR-2 inhibitor, both have vasodilatory effects. The combination of these drugs could lead to an increased risk of hypotension due to additive vasodilation. Reference examples such as bosentan increasing the hypotensive activities of penbutolol support this interaction. Therefore, the most plausible interaction type is that riociguat may increase the hypotensive activities of macitentan, leading to a higher risk of fainting, falls, or shock. Close clinical monitoring and adjustment of dosages may be necessary to manage this risk. 
  \\\hline
  Pharmacokineticist \newline (Iteration 1)& \textbf{Assigned knowledge sources:} Absorption DDIs, Metabolic DDIs and Synergistic DDIs.  \newline \textbf{Answer: Interaction type D.} Riociguat is known to increase cGMP levels by inhibiting PDE5, which can lead to vasodilation. Macitentan, on the other hand, is an endothelin receptor antagonist that also has vasodilatory properties. The combination of these two drugs, both of which have vasodilatory effects, could potentially lead to a significant increase in blood flow, resulting in elevated blood levels of Macitentan. This could increase both the efficacy and the toxicity risk of Macitentan, given its established side effects including hypotension and cardiovascular events. Therefore, the most plausible interaction is that Riociguat increases the serum concentration of Macitentan, potentially leading to higher efficacy and toxicity risk.
  \\\hline
  Pharmacologist \newline (Iteration 1)& \textbf{Assigned knowledge sources:} Metabolic DDIs, Antagonistic DDIs, Drug targets and Drug side effects.  \newline \textbf{Answer: Interaction type E.} Riociguat increases cyclic guanosine monophosphate (cGMP) levels through the inhibition of phosphodiesterase type 5 (PDE5), leading to vasodilation. Macitentan, on the other hand, is a VEGFR-1 and VEGFR-2 inhibitor, which is not directly involved in the PDE5 pathway. However, the mechanism of action of both drugs involves cardiovascular effects. Riociguat is primarily used to treat pulmonary arterial hypertension, while Macitentan is used to treat pulmonary arterial hypertension and systemic sclerosis. The combination of these two drugs could potentially lead to an increased hypotensive effect due to the synergistic vasodilation, which aligns with the description of interaction type E. This interaction type suggests that riociguat would enhance the hypotensive activities of macitentan, increasing the risk of excessive blood pressure reduction and related adverse effects such as dizziness, fainting, or shock. Therefore, this is the most plausible interaction type based on the provided information and medical knowledge.
  \\\hline
  Conclusion agent \newline (Iteration 1)&  \textbf{No.} \textbf{Pharmacist:} Interaction type E is suggested by multiple experts, indicating an increased hypotensive effect. \newline \textbf{Pharmacokineticist:} Interaction type D is proposed, indicating an increased serum concentration of Macitentan. \newline \textbf{Pharmacologist:} Interaction type E is favored, suggesting an increased risk of adverse effects due to synergistic vasodilation. \newline Further analysis is needed to clarify the most likely interaction type. A more detailed review of clinical studies and case reports involving these drugs in combination is recommended. Additionally, pharmacokinetic studies to assess the impact of riociguat on the metabolism or absorption of macitentan could provide more concrete evidence.
  \\\hline
  Planner agent for \newline expert agent \newline instantiation \newline (Iteration 2)&  \textbf{Cardiologist:} provide insights into the cardiovascular effects and the potential for increased hypotension. \newline \textbf{Pulmonologist:} offer expertise on the respiratory effects and the interaction between these drugs, given their role in treating pulmonary arterial hypertension. \newline \textbf{Nephrologist:} assess the potential renal implications of drug combination.
  \\\hline
  Cardiologist \newline (Iteration 2)& \textbf{Assigned knowledge sources:} Absorption DDIs, Metabolic DDIs, Antagonistic DDIs and Drug targets.  \newline  \textbf{Answer: Interaction type E.} Riociguat and Macitentan both have vasodilatory effects. Riociguat increases cyclic guanosine monophosphate (cGMP) levels, leading to vasodilation, while Macitentan inhibits VEGFR-1 and VEGFR-2, which also have vasodilatory effects. Given the expert opinions and the deep learning model predictions, the most plausible interaction is that Riociguat may increase the hypotensive activities of Macitentan. This interaction would result in a higher risk of excessive hypotension, dizziness, fainting, or falls. The clinical monitoring and titration recommended by experts align with this hypothesis. Therefore, interaction type E is the most plausible based on the provided information and medical knowledge.
  \\\hline
  \end{tabular}
  \label{app:example2}
\end{table*}

\begin{table*}[h]
  \centering
  \caption{A DDI prediction case by DDIAgents that different agents achieve concensus in two iterations (continued).}
  \vspace{-5px}
  \begin{tabular}{p{2.5cm}|p{14.5cm}}
  \hline
  Pulmonologist \newline (Iteration 2)& \textbf{Assigned knowledge sources:} Absorption DDIs, Metabolic DDIs, Drug targets and Drug side effects.  \newline \textbf{Answer: Interaction type D.} Based on the provided descriptions and the deep learning model's predictions, there are indications that riociguat may enhance the hypotensive effects of macitentan, which is a significant concern given the clinical context of pulmonary arterial hypertension (PAH) where maintaining adequate blood pressure is critical. The pharmacologist's suggestion of interaction type E, which indicates an increased risk of adverse effects due to synergistic vasodilation, aligns with the known mechanisms of action for both drugs. Riociguat works by increasing nitric oxide bioavailability and cGMP levels, leading to vasodilation, while macitentan inhibits VEGFR-1 and VEGFR-2, also promoting vasodilation. The combination could result in excessive vasodilation, leading to hypotension. This is further supported by the pharmacist's and pharmacologist's consensus favoring interaction type E. The deep learning model's prediction of an increased hypotensive activity is the most plausible based on the available information and expert opinions.
  \\\hline
  Nephrologist \newline (Iteration 2)& \textbf{Assigned knowledge sources:} Absorption DDIs, Metabolic DDIs, Excretion DDIs and Synergistic DDIs.  \newline \textbf{Answer: Interaction type E.} Riociguat is known to cause vasodilation, and Macitentan also has vasodilatory effects by inhibiting VEGFR-1 and VEGFR-2. Given that both drugs can lead to hypotension, the combination is likely to enhance this effect. The deep learning model suggests an increased hypotensive activity, which aligns with the pharmacological actions of both drugs. The pharmacist and pharmacologist have also favored this interaction type, indicating a consensus on the potential for increased hypotension. Therefore, the most plausible interaction type is that riociguat may increase the hypotensive activities of Macitentan, leading to a heightened risk of fainting, falls, or shock.
  \\\hline
  Conclusion agent \newline (Iteration 2)&  \textbf{Yes. Interaction type E.} Based on the comprehensive analysis of the provided information and expert opinions, the most plausible interaction between Riociguat and Macitentan is that Riociguat may increase the hypotensive activities of Macitentan. Both drugs have vasodilatory effects, with Riociguat increasing cyclic guanosine monophosphate (cGMP) levels and Macitentan inhibiting VEGFR-1 and VEGFR-2. The deep learning model also indicates an increased hypotensive activity, which aligns with the known mechanisms of action. Furthermore, the consensus among the pharmacologist, pharmacist, and other experts supports this interaction type, emphasizing the potential risk to explain. 
  \\\hline
  \end{tabular}
  \label{app:example2new}
\end{table*}

\subsection{An Illustrative Example for Comparison on Generation Contents}
\label{app:gencont}
In order to provide a clearer comparison between the generated contents of MDAgents and DDIAgents in Section~\ref{exp:interpertability}, we present an illustrative example in Table~\ref{tab:gen_comp}. 
The contents include the DDI question, the answer by MDAgents and DDIAgents, and the LLM judgment. 
We can see that DDIAgents outperforms in all of the evaluation dimensions including thoroughness, diversity, rationality and overall quality. 

\begin{table*}[h]
  \centering
  \caption{Illustrative case: generated content comparison between MDAgents and DDIAgents.}
  \vspace{-5px}
  \begin{tabular}{p{2cm}|p{15cm}}
  \hline
  DDI prediction question& What is the drug-drug interaction type between Aclidinium and Fesoterodine? \newline A. Increased risk or severity of adverse effects B. Increased anticholinergic activities \newline C. Increased tachycardic activities D. Increased the QTc-prolonging activities E. Increased serum concentration \\\hline
  MDAgents& Interaction type A. The evidence provided from the deep learning model and the analysis of the agents strongly supports the interaction type A, where Aclidinium may increase the anticholinergic activities of Fesoterodine. The detailed explanations from the agents highlighting the anticholinergic effects of both drugs, provides a robust basis for this conclusion. \\\hline
  DDIAgents& Interaction type B. Both aclidinium and fesoterodine are anticholinergic agents, and their concurrent use can potentiate anticholinergic side effects, making interaction type B (increased anticholinergic activities) the most precise and clinically relevant interaction type. This is supported by the analysis provided by the pharmacists and pharmacologists, who emphasize the pharmacodynamic overlap and the potential for enhanced side effects when these drugs are used together. \\\hline
  LLM Decision&Thoroughness: "Winner": Answer 2~(DDIAgents). \newline Explanation: Answer 2 is more comprehensive because it accurately aligns the drug mechanism with the specific multiple-choice option provided. It correctly identifies both drugs as anticholinergic agents and links this to Interaction type B. Answer 1, conversely, suffers from a mapping error; it selects "Interaction type A" but then proceeds to describe the definition of "Interaction type B" in its explanation, creating a contradiction that fails to fully address the question's structure. \newline \newline Diversity: "Winner": Answer 2~(DDIAgents). \newline Explanation: Answer 2 offers a richer clinical perspective. It mentions "pharmacodynamic overlap" and the perspectives of "pharmacists and pharmacologists," which provides a broader professional context for why this interaction is significant. Answer 1 mentions a "deep learning model," which is a technical insight, but it lacks the clinical depth found in Answer 2. \newline \newline Rationality: "Winner": Answer 2~(DDIAgents). \newline Explanation: Answer 2 is logically sound. It identifies the shared drug class and correctly matches it to the most specific answer choice (B). Answer 1 is irrational in its presentation: it claims the answer is "A" (the general risk of adverse effects) but then justifies it using the exact phrasing of option "B" (anticholinergic activities). This internal inconsistency makes Answer 1 unreliable. \newline \newline Overall: "Winner": Answer 2~(DDIAgents). \newline  Explanation: Answer 2 is the superior response. It is factually accurate, logically consistent, and maps the pharmacological mechanism to the correct multiple-choice option. Answer 1 contains a critical labeling error—matching the explanation for B to the label for A—which would lead a user to the wrong choice despite having the correct underlying reasoning. Answer 2 provides the "most precise and clinically relevant" answer, as it correctly identifies the specific nature of the interaction. \\\hline
  \end{tabular}
  \label{tab:gen_comp}
\end{table*}

\clearpage

\end{document}